\documentclass[sn-mathphys, table]{sn-jnl}	
\usepackage{subcaption}
\usepackage{bm}
\usepackage{rotating}
\usepackage{algorithmicx}
\usepackage{makecell}

\usepackage{pifont}
\newcommand{\cmark}{\ding{51}}%
\newcommand{\xmark}{\ding{55}}%
\usepackage{cancel}
\usepackage{xcolor}
\usepackage{soul}

\newcommand{\minus}{\scalebox{0.75}[1.0]{$-$}}

\graphicspath{{Plots/}{/Users/johnmartin/Documents/Research/ML_Gravity/Plots/}}

\begin{document}

\title[The Physics-Informed Neural Network Gravity Model Generation III]{The Physics-Informed Neural Network Gravity Model Generation III}

\author*[1]{\fnm{John} \sur{Martin}}\email{jrmartin@umd.edu}

\author[2]{\fnm{Hanspeter} \sur{Schaub}}\email{hanspeter.schaub@colorado.edu}

\affil*[1]{\orgdiv{Department of Aerospace Engineering}, \orgname{University of Maryland}, \orgaddress{\street{4298 Campus Dr}, \city{College Park}, \postcode{20742}, \state{MD}, \country{USA}}}
\affil[2]{\orgdiv{Ann and H. J. Smead Department of Aerospace Engineering Sciences}, \orgname{University of Colorado Boulder}, \orgaddress{\street{431 UCB}, \city{Boulder}, \postcode{80309}, \state{CO}, \country{USA}}}

\abstract{Scientific machine learning and the advent of the Physics-Informed Neural Network (PINN) have shown high potential in their ability to solve complex differential equations. One example is the use of PINNs to solve the gravity field modeling problem --- learning convenient representations of the gravitational potential from position and acceleration data. These PINN gravity models, or PINN-GMs, have demonstrated advantages in model compactness, robustness to noise, and sample efficiency when compared to popular alternatives; however, further investigation has revealed various failure modes for these and other machine learning gravity models which this manuscript aims to address. Specifically, this paper introduces the third generation Physics-Informed Neural Network Gravity Model (PINN-GM-III) which includes design changes that solve the problems of feature divergence, bias towards low-altitude samples, numerical instability, and extrapolation error. Six evaluation metrics are proposed to expose these past pitfalls and illustrate the PINN-GM-III's robustness to them. This study concludes by evaluating the PINN-GM-III modeling accuracy on a heterogeneous density asteroid, and comparing its performance to other analytic and machine learning gravity models.} 


\maketitle

\section{Introduction}
\label{sec:introduction}

Nearly all problems in astrodynamics involve the force of gravity. Be it in trajectory optimization, spacecraft rendezvous, orbit determination, or other problems, gravity often plays a significant --- if not dominant --- role in the system dynamics. The ubiquity of this force is a testament to its significance, yet despite this, there exist surprisingly few ways to represent this force to high accuracy. The construction of high-fidelity gravity models is henceforth referred to as the gravity modeling problem, and currently there exists no universally adopted solution. Some gravity models perform well for large planetary bodies, but break down when modeling objects with exotic geometries like asteroids and comets. Other models can handle irregular shapes but require assumptions and come with high computational cost. The one commonality is that all existing gravity models come with their own unique pitfalls that prevent standardization across the community. Consequently, researchers are continuing to design new solutions to the gravity modeling problem with hopes of one day finding a universal model. One encouraging vein of research explores the use of neural networks and scientific machine learning to learn gravity models free of these past difficulties. 

\begin{figure}[h!tb]
    \centering\includegraphics[width=\textwidth]{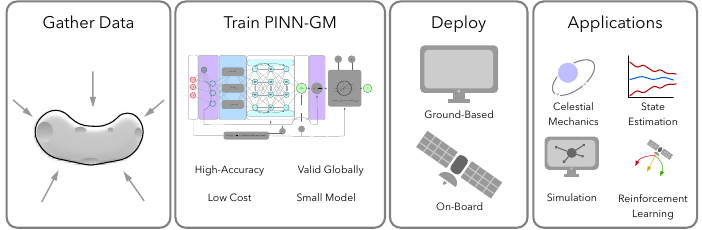}
    \caption{PINN-GMs offer high-accuracy, low-cost gravity solutions to be deployed across a variety of applications.}
    \label{fig:concept}
\end{figure}

Scientific machine learning uses neural networks to generate high-fidelity models of, and solutions to, complex differential equations~\cite{cuomoScientificMachineLearning2022, karniadakisPhysicsinformedMachineLearning2021}. Physics-informed neural networks (PINNs) are one class of model available to solve such problems. PINNs are neural networks trained in a manner that intrinsically respects relevant differential and physics-based constraints. This compliance, or ``physics-informing'', is achieved by augmenting the network cost function with said constraints, such that the learned model is penalized for violating any of the underlying physics. Through this design change, PINNs have been shown to achieve higher accuracy with less data than their traditional neural network counterparts~\cite{raissiPhysicsinformedNeuralNetworks2019}. 

Recently, researchers have proposed the use of PINNs as well as other machine learning models to solve the gravity modeling problem~\cite{martinEarthAndMoon2022, martinSmallBodies2022}. These models have been shown to produce high-accuracy solutions in both large- and small-body settings under specific training conditions. While these models demonstrate early promise, closer inspection reveals that there remain regimes in which these machine learning models perform unreliably. This paper exposes these shortcomings and proposes solutions to them through multiple design changes to the underlying machine learning architecture. Specifically, this paper introduces the third generation of the PINN gravity model, or PINN-GM, which includes design changes that solve the problems of feature divergence, extrapolation error, numerical instability, bias towards low-altitude samples, and incompliant boundary conditions.

\section{Background}
\label{sec:methods}

Gravity field models can be categorized into two groups: analytic and numerical. Analytic models are derived from first principles and yield closed-form equations for the gravitational potential, whereas numerical models are constructed in a data-driven manner that come without an explicit equation. Each model has its own set of advantages and drawbacks, and the choice of which model to use is often dictated by application. The following section briefly surveys the available gravity models and their corresponding pros and cons.

\subsection{Analytic Models}

\subsubsection*{Spherical Harmonics Model}

In the 1900s, spherical harmonic basis functions were proposed to represent high-order perturbations in the Earth's gravity field~\cite{brillouinEquationsAuxDerivees1933}. These harmonics can be superimposed to produce the spherical harmonic gravity model through:
\begin{equation}
    U(r) = \frac{\mu}{r} \sum_{l=0}^l \sum_{m=0}^l \bigg(\frac{R}{r}\bigg)^l P_{l,m}\left( \sin \phi \right) \big[C_{l,m} \cos(m \lambda) + S_{l,m} \sin(m \lambda)\big]
    \label{eq:spherical_harmonics}
\end{equation}
where $r$ is the distance to the test point, $\mu$ is the gravitational parameter of the body, $R$ is the circumscribing radius, $l$ and $m$ are the degree and order of the model, $C_{l,m}$ and $S_{l,m}$ are the Stokes coefficients, $\lambda$ and $\phi$ are the longitude and latitude, and $P_{l,m}$ are the associated Legendre polynomials~\cite{kaulaTheorySatelliteGeodesy1966}. The spherical harmonic gravity model is commonly used to represent the fields of large planetary bodies like the Earth~\cite{pavlisPreliminaryGravitationalModel2005}, the Moon~\cite{goossensGlobalDegreeOrder2016}, and Mars~\cite{genovaSeasonalStaticGravity2016a}, as they are among the most efficient models for capturing planetary oblateness --- the largest gravitational perturbation found on these large, rotating bodies. Using only a single coefficient, $C_{2,0}$, spherical harmonics can succinctly capture this important gravitational feature and its effects on spacecraft motion. 

While these models are effective at representing planetary oblateness, they struggle to model the remaining gravitational perturbations like mountain ranges, tectonic plate boundaries, and craters. These discontinuous features are notoriously hard to represent using periodic basis functions --- requiring the superposition of hundreds of thousands of high-frequency harmonics to overcome the 3D equivalent of Gibbs phenomenon~\cite{hewittGibbsWilbrahamPhenomenonEpisode1979}. These higher frequencies introduce an $\mathcal{O}(n^2)$ computational and memory cost~\cite{martinGPGPUImplementationPines2020}, and the regression of these harmonics is especially difficult due to their rapidly fading observability. High-order spherical harmonic models also can diverge when evaluated within the sphere that bounds all mass elements (Brillouin sphere). While these effects are negligible for near-spherical planets or moons, they can become problematic in small-body settings where objects can exhibit highly non-spherical geometries~\cite{wernerExteriorGravitationPolyhedron1997}.  

\subsubsection*{Polyhedral Gravity Model}

The polyhedral gravity model is a popular alternative in these small-body settings, offering a solution that maintains validity down to the surface of any object regardless of shape. These models require a preexisting shape model of the asteroid or comet --- a collection of triangular facets and vertices which captures its geometry --- from which an analytic acceleration can be computed through:
\begin{equation}
    \nabla U=-G \sigma \sum_{e \in \text{edges}} \mathbf{E}_{e} \cdot \mathbf{r}_{e} \cdot L_{e}+G \sigma \sum_{f \in \text{facets}} \mathbf{F}_{f} \cdot \mathbf{r}_{f} \cdot \omega_{f}
    \label{eq:polyhedral}
\end{equation}
where $G$ is the gravitational constant, $\sigma$ is the density of the body, $\mathbf{E}_e$ is an edge dyad, $\mathbf{r}_e$ is the position vector between the center of the edge and the test point, $L_e$ is an analog to the potential contribution of the edge, $\mathbf{F}_f$ is the face normal dyad, $\mathbf{r}_f$ is the distance between the face normal and the test point, and $\omega_f$ is an analog to the potential contribution of the face~\cite{wernerExteriorGravitationPolyhedron1997}. 

While the polyhedral gravity model avoids divergence within the Brillouin sphere, it comes with its own challenges. First, this gravity model is expensive when evaluated on high-resolution shape models --- requiring intensive summation loops over all vertices and facets to compute the acceleration at a single test point. The model also assumes that the body's density is known a priori. While a constant density assumption is often used, literature has shown that such assumption is not necessarily valid~\cite{scheeresEstimatingAsteroidDensity2000,scheeresHeterogeneousMassDistribution2020}. 

\subsubsection*{Mascon Gravity Model}

In contrast to the analytically derived spherical harmonic and polyhedral models, the mascon gravity models instead approximates the gravitational potential using a collection of point mass elements, known as mascons~\cite{mullerMasconsLunarMass1968}. These masses are distributed within the body and summed to form a global approximation of the gravity field through Equation~\ref{eq:mascons}:
\begin{equation}
    \bm{a}(\bm{r}) = \sum_{k=0}^N \mu_k\frac{\bm{r} - \bm{r}_k}{\lVert \bm{r} - \bm{r}_k \rVert^3}
    \label{eq:mascons}
\end{equation}
where $\mu_k$ and $\bm{r}_k$ are the estimated gravitational parameter and position for the $k$-th mascon.

While the mascon model offers an efficient alternative to the polyhedral model, it becomes inaccurate when evaluated near individual mascons~\cite{tardivelLimitsMasconsApproximation2016}. Hybrid mascon models offer a more accurate alternative --- representing each mascon with a low fidelity spherical harmonic model --- but this incorporates additional complexity in their regression and remains prone to the same challenges of the traditional mascon and spherical harmonic models~\cite{wittickMixedmodelGravityRepresentations2019}.

\subsubsection*{Other Analytic Models}
The ellipsoidal harmonic model follows a similar approach to that of spherical harmonics but uses ellipsoidal harmonic basis functions instead~\cite{romainEllipsoidalHarmonicExpansions2001}. This yields a smaller region in which the model could diverge; however, it still struggles to model discontinuity with its periodic basis functions. The interior spherical harmonic model inverts the spherical harmonic formulation and can model a local region whose boundary intersects only one point on the surface of the body~\cite{takahashiSurfaceGravityFields2013}. This model maintains stability down to that single point making it valuable for precise landing operations; however, the solution becomes invalid on any other point on the surface and outside of the corresponding local sphere. Finally, the interior spherical Bessel gravity model expands upon the interior model but uses Bessel functions rather than spherical harmonics to achieve a wider region of validity. This model comes with added analytic complexity and can also struggle to capture discontinuous features efficiently~\cite{takahashiSmallBodySurface2014}.

\subsection{Machine Learning Models}

As an alternative to analytic gravity models, recent efforts explore the use of machine learning to regress models of complex gravity fields in a data-driven manner. Unlike analytic approaches, machine learning models are generally free of assumptions about the body they are modeling and have no analytic limitations --- e.g. divergence in the bounding sphere, required shape models, etc. These models have historically required large volumes of training data; however once trained, they can offer high-accuracy predictions at comparatively low computational cost. Table~\ref{tab:ML_Gravity} summarizes key accuracy and training metrics for recently reported machine learning gravity models, and a specific discussion for each model is provided below.  

\begin{table}[h!]
 \centering
  \begin{tabular}{l r r r r} 
    \toprule
    Model &  Parameters & Training Data & Avg. Error [\%] & Valid Globally \\ [0.5ex] 
    \midrule
    GP~\cite{gaoEfficientGravityField2019} & 12,960,000 & 3,600 & 1.5\% & \xmark \\ 
   NNs~\cite{chengRealtimeControlFueloptimal2019} &  1,300,000 & 800,000 & 0.35\% & \xmark \\
   ELMs~\cite{furfaroModelingIrregularSmall2020} & 350,000 & 768,000 & 1-10\% & \xmark \\
   GeodesyNet~\cite{izzoGeodesyIrregularSmall2022} & 80,800 & 1,000,000 & 0.36\% & \cmark \\
   PINN-GM-III & 2,211 & 4,096 & 0.30\% & \cmark \\ [1ex] 
   \bottomrule
  \end{tabular}
  \caption{Machine Learning Gravity Model Statistics -- See Appendix~\ref{app:ml_comparision}}
  \label{tab:ML_Gravity}
\end{table}

\subsubsection*{Gaussian Processes}
Gaussian processes, or GPs, are non-parametric models that are fit by specifying a prior distribution over functions, and updating that prior based on observed data. This requires the user to first specify some kernel function which measures similarity between data and then compute a covariance matrix between all data pairs using that function. Once computed, the covariance matrix is inverted and used to evaluate the mean and uncertainty of the learned function at a test point.

GPs were proposed to solve the gravity modeling problem in 2019~\cite{gaoEfficientGravityField2019}, regressing a mapping between position and acceleration training data. GPs are advantageous because they provide a probabilistic estimate of the uncertainty in the model's prediction; however these models do not scale well to large data sets. The GP's covariance matrix is built from the training data and scales as $\mathcal{O}(n^2)$, where $n$ is the size of the training data set, and the computational complexity of the matrix inversion scales as $\mathcal{O}(n^3)$. This scaling makes it impractical to fit GPs using large quantities of data, intrinsically limiting its utility and performance. 

\subsubsection*{Extreme Learning Machines}
Extreme learning machines (ELMs) have also been proposed to predict gravitational accelerations~\cite{furfaroModelingIrregularSmall2020}. ELMs are single layer neural networks fit by randomly initializing the weights from the inputs to the hidden layer, and then computing the optimal weights to the output layer using least-squares regression. Explicitly the ELM minimizes the mean-square error loss function:
\begin{equation}
    L(\bm{\theta}) = \frac{1}{N}\sum_{i=0}^N \lVert \hat{\bm{y}}_i(\bm{x}_i\vert \bm{\theta}) - \bm{y}_i \rVert ^2
    \label{eq:traditional_loss}
\end{equation}
where $\hat{y}_i(\bm{x_i}\vert \bm{\theta})$ is the machine learning model prediction at input $\bm{x}_i$ with trainable model parameters $\bm{\theta}$~\cite{huangExtremeLearningMachine2006}. 

ELMs are advantageous because they can model non-linear functions, and they only require a single training iteration. However, depending on the width of the ELM and the amount of training data used, these models can be prone to memory and computational difficulties due to the large $\mathcal{O}(n^3)$ matrix inversion used in the least squares solution. Iterative chunking strategies have been proposed to remedy this issue, though this makes the relative advantage between ELMs and traditional neural networks less apparent. Moreover, these models are data intensive, such that past solutions have required hundreds of thousands of training data to produce high-accuracy models. 

\subsubsection*{Neural Networks}

Neural networks are similar to ELMs, but they are typically multi-layer and fit using gradient decent and backpropagation rather than least squares. These models iteratively update their parameters using small batches of training data combined with various optimization algorithms~\cite{kingmaAdamMethodStochastic2014, dozatIncorporatingNesterovMomentum, nocedalNumericalOptimization2006}. Neural networks have also been proposed as a candidate gravity model~\cite{chengRealtimeOptimalControl2020}. When juxtaposed with ELMs, neural networks' primary benefit is their lack of required matrix inverse; however, this comes at the cost of iterative training which can take long periods of time. Neural networks share similar drawbacks to ELMs, requiring large quantities of data and are prone to extrapolation error when evaluated outside of the bounds of the training data.

\subsubsection*{Physics-Informed Neural Networks}

Physics-informed neural networks, or PINNs, are yet another model proposed to solve the gravity modeling problem~\cite{martinEarthAndMoon2022,martinSmallBodies2022}. These physics-informed models increase sample efficiency over traditional networks by incorporating differential constraints into the loss function of the neural network. These constraints limit the set of learnable functions to only those that comply with the underlying physics~\cite{raissiPhysicsinformedNeuralNetworks2019}. The first PINN gravity model, or PINN-GM, uses the known differential equation $\minus \nabla U = \bm{a}$ to define the loss function:  
\begin{equation}
    L(\theta) = \frac{1}{N}\sum_{i=0}^N \lVert \minus \nabla \hat{U}(x_i\vert \theta) - \bm{a}_i  \rVert ^2
    \label{eq:pinn_I_loss}
\end{equation} 
where $\hat{U}$ is the potential learned by the network, which can then be differentiated via automatic differentiation~\cite{baydinAutomaticDifferentiationMachine2018} to compute the corresponding acceleration. The inclusion of these physics constraints, combined with network design modifications, have shown that PINN-GMs can yield solutions that maintain similar accuracy to their predecessors while using orders of magnitude fewer parameters and data. These models have also demonstrated enhanced robustness to uncertainty in the training data as a result of their physics-informed constraints~\cite{martinSmallBodies2022}.

\subsubsection*{GeodesyNet}
Finally, in 2022 Reference~\citenum{izzoGeodesyIrregularSmall2022} introduced GeodesyNets as a candidate solution to the gravity modeling problem. GeodesyNets are neural density fields --- close relatives to the popular neural radiance fields~\cite{mildenhallNeRFRepresentingScenes2022} --- which use neural networks to learn a density function for every point in a 3D volume. Once learned, these density fields can be numerically integrated to yield a gravitational acceleration or a value of the potential. This work is an exciting new class of machine learning model capable of achieving high accuracy given sufficient quantities of data. These models also behave more reliably at high-altitudes as the networks are not required to predict densities beyond a unit volume; albeit the cost required to perform the numerical integration can be relatively high --- requiring evaluation of 30,000 to 300,000 quadrature points per prediction~\cite{izzoGeodesyIrregularSmall2022} --- which can make these models cumbersome to evaluate and train. Finally, a key feature of these models is their ability to estimate an internal density profile of an asteroid --- a capability presumed available to PINNs by evaluating Poisson's equation inside the body --- but such capabilities are not the focus of this work.

\subsection{Current Challenges}

As these machine learning gravity models become more mainstream, they warrant further scrutiny. While these models can yield accurate solutions under ideal training and test cases, further investigation reveal that these models are not yet universally robust. Rather, there exist common problems that exist for the majority of these models that have yet to be identified or addressed. Specifically, these models are prone to extrapolation error and numerical instabilities, and they are also relatively brittle to sparse and noisy data conditions. If the community strives to develop a universally valid machine learning gravity model, these challenges need to be addressed. 

In this manuscript, we aim to expose these various failure cases and introduce design modifications to the underlying machine learning architectures that can increase the their robustness and generalizability. To accomplish this, we introduce the third generation PINN gravity model, or PINN-GM-III, which includes a variety of design changes that improve modeling accuracy and robustness across a wide set of training and test cases. These modifications, and the failure modes they eliminate, are discussed at length in Section 3. While these modifications are applied specifically to the PINN-GM, it should be noted that some can be applied to other machine learning gravity models as well. Details of the specific PINN modifications and their impact on performance are supplied below. 

\section{PINN-GM-III}
\label{sec:pinn_gm}

The training process for the original PINN-GM can be found in Ref.~\cite{martinSmallBodies2022} and is briefly summarized here for convenience. To begin, the neural network requires a set of $N$ position and acceleration vectors sampled around a celestial body:
\begin{equation}
    \bar{\bm{x}} = \begin{bmatrix}
        \bm{x}_1 \\
        \bm{x}_2 \\
        \vdots \\
        \bm{x}_N 
    \end{bmatrix}
    \quad
    \bar{\bm{a}} = \begin{bmatrix}
        \bm{a}_1 \\
        \bm{a}_2 \\
        \vdots \\
        \bm{a}_N 
    \end{bmatrix}
\end{equation}
This data can be collected in one of two ways: 1) It can be generated synthetically from pre-existing high-fidelity gravity models like EGM-2008 for Earth~\cite{pavlisDevelopmentEvaluationEarth2012} or a high-fidelity polyhedral model for asteroids~\cite{gaskellGaskellErosShape2021}, or 2) the data can be estimated on-board without any prior knowledge of the field. The latter can be accomplished using estimation techniques like dynamic model compensation~\cite{leonardGravityErrorCompensation2013} to simultaneously estimate spacecraft position and accelerations in-situ as is shown in Ref.~\cite{martinPINNFilter2022}. For this study, all data will be synthetically generated and evaluation on real data is left for future work.

\begin{figure}[htb]
    \centering\includegraphics[width=\textwidth]{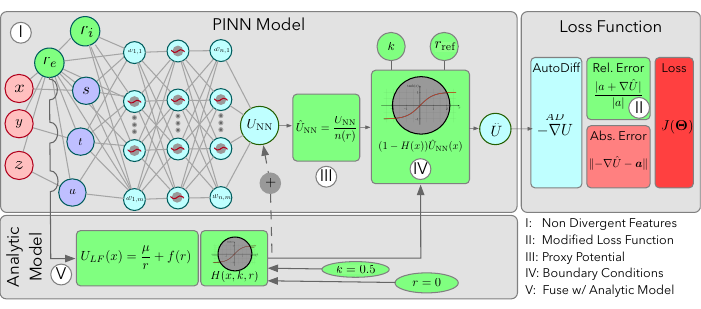}
    \caption{PINN-GM Generation III with new modifications highlighted in green}
    \label{fig:PINN-GM-III}
\end{figure}

Once the data is gathered, a neural network is fit to learn a mapping between these data. The neural network takes the form: 
\begin{equation}
\hat{y}_{\bm{\theta}}(\bm{x}) = W^{(L)} \cdot \sigma(W^{(L-1)} \cdot \ldots \cdot \sigma(W^{(1)} \cdot x + b^{(1)}) + b^{(L-1)}) + b^{(L)}
\end{equation}
where $\hat{y}_{\bm{\theta}}$ is the output of the neural network parameterized by weights and biases ${{W}, {b}} \in \bm{\theta}$. These parameters are found in the 0th to L-th hidden layers of the network. Successive non-linear transformations of the network input $\bm{x}$ are applied via the activation function, $\sigma$, to yield $\hat{y}_{\bm{\theta}}(\bm{x})$. 

For PINN-GMs, the output of the neural network corresponds to the predicted gravitational potential, $\hat{U}$, at position $\bm{x}$. That potential is then differentiated to produce an acceleration vector which is used in the network's physics-informed loss function: 
\begin{equation}
L(\bm{\theta}) = \frac{1}{N}\sum_{i=0}^N \lVert \minus \nabla \hat{U}(\bm{x_i}\vert \bm{\theta}) - \bm{a}_i  \rVert ^2
\end{equation}
The network is iteratively trained using stochastic gradient descent which estimates the gradient of the loss with respect the network parameters, $\nabla_{\bm{\theta}}L$. For every iteration, the parameters are updated via:
\begin{equation}
\bm{\theta}' \rightarrow \bm{\theta} - \eta \nabla_{\bm{\theta}}L
\end{equation}
where $\eta$ is a user-defined learning rate. 

The second generation PINN-GM, or PINN-GM-II, enhances this training process by augmenting the loss function with additional constraints to increase robustness to noise. In addition, better input features are selected to increase sample efficiency. Despite these improvements, closer inspection has revealed that the PINN-GM-II has failure cases that remain unaddressed. While these pitfalls are explicitly exposed within the PINN-GM-II, they are general to other published machine learning gravity models to the best of the authors' knowledge. The subsequent sections outline these pitfalls and propose architectural changes to the underlying model architecture to address them. Collectively, these changes form the third generation PINN gravity model, or PINN-GM-III. The modifications are illustrated in Figure~\ref{fig:PINN-GM-III} and discussed in detail in Sections~\ref{sec:preprocessing} through \ref{sec:analytic_model}, with their combined effect demonstrated in Section~\ref{sec:visualization_of_mods}.

\begin{algorithm}
\caption{PINN-GM-III algorithm}
\small
\begin{algorithmic}[1]
\State Collect training data ($\bm{x}, \bm{a}$) from: \newline
    (a) a pre-existing model \newline
    (b) online state estimates~\cite{martinPINNFilter2022}
\State Non-dimensionalize the training data \Comment{App.~\ref{app:training_details}}
\State Convert to 5D non-singular spherical coordinates ($r_i$, $r_e$, $s, t, u$) \Comment{Sec.~\ref{sec:preprocessing}}
\State Propagate through the neural network
\State Output proxy potential $U_{\text{NN}}$ \Comment{Sec.~\ref{sec:proxy_potential}}
\State Scale proxy potential into true potential $\hat{U}_{\text{NN}}$ \Comment{Sec.~\ref{sec:proxy_potential}}
\State \textit{(Optional)} Fuse with weighted low-fidelity potential $\hat{U}_{\text{LF}}$ \Comment{Sec.~\ref{sec:analytic_model}} 
\State Enforce boundary conditions on the network potential $\hat{U}$  \Comment{Sec.~\ref{sec:boundary_conditions}} 
\State Autodifferentiate (AD) potential to produce acceleration $\hat{\bm{a}}$
\If{Training}
\State Sum acceleration percent error and absolute error to form loss  \Comment{Sec.~\ref{sec:loss_function}}
\State Compute gradients of loss function
\State Update network parameters
\EndIf
\end{algorithmic}
\end{algorithm}

\subsection{Feature Engineering}
\label{sec:preprocessing}
The first pitfall of past machine learning gravity models is the choice of inefficient and divergent input features. Cartesian position coordinates are most regularly used as inputs for these models; however, these coordinates are prone to two major drawbacks. First, they span the set of all real numbers, and the networks require data sampled across this entire domain to learn robust solutions for any possible value of $\bm{x}$. The PINN-GM-II first identified this problem and proposed a conversion to 4D non-singular spherical coordinates $(r, s, t, u)$ where $s,t,$ and $u$ are the sine of the angle between the test point and the cartesian axes. While the domain for $s,t,$ and $u$ was reduced to $[-1,1]$ which increases the model's sample efficiency, the radial coordinate, $r$, remains prone to the second major drawback: feature divergence. 

When evaluating these numerical models at points far from the body, the radial coordinate remains unbounded and can introduce numerical instabilities that cause the model to diverge. While the radial coordinate can be converted to $1/r$, this instead introduces instability for test points near the surface of the body. In both cases, the radial feature can have magnitudes greater than one which will cause many activation functions to prematurely saturate and decrease the learning efficiency of the model~\cite{goodfellowDeepLearning2016}.   

This pitfall can be addressed through a simple design modification where the Cartesian position coordinates are converted into a 5D spherical coordinate description of $(r_i, r_e, s, t, u)$ where $r_e$ and $r_i$ are two proxies of the test point radius, $r$, defined as \begin{equation}
    r_i = \begin{cases}
        r &\quad r \in [0,R] \\
        1 &\quad r \in [R, \infty)
    \end{cases}
    \quad \text{and} \quad 
    r_e = \begin{cases}
        1 &\quad r \in [0,R] \\
        \frac{1}{r} &\quad r \in [R, \infty)
    \end{cases} 
\end{equation}
Using this convention, the network will maintain the desired sample efficiency of past approaches while also ensuring that its input features can never diverge regardless of the location of any training or test point. This modification constitutes the first design change of the PINN-GM-III.

\subsection{Modified Loss Function to Account for High-Altitude Samples}
\label{sec:loss_function}
The second pitfall of past machine learning gravity models arises from their default loss functions which inadvertently decrease modeling accuracy at high-altitudes. Specifically, most machine learning gravity models use an absolute error loss function --- e.g. mean squared error (MSE), root mean squared error (RMS). For the original PINNs, this was captured through the norm of the differenced acceleration vectors:
\begin{equation}
    \mathcal{L}_{\text{ABS}}(\mathbf{\theta}) = \frac{1}{N}\sum^{N}_{i=0}\left\lVert \minus \nabla \hat{U}(\bm{x}_i\vert \mathbf{\theta}) - \bm{a}_i\right\rVert
    \label{eq:L_RMS}
\end{equation}
While these loss functions will successfully minimize the most flagrant residuals between the predicted and true acceleration vectors; they induce an undesirable effect at high-altitudes. Because accelerations produced near the surface of a celestial body have much larger magnitudes than the accelerations produced at high-altitudes, low-altitude errors will always appear disproportionately large compared to any errors at high-altitude. As a consequence, loss functions that minimize absolute errors will always prioritize low-altitudes samples, even if the high-altitude predictions are more erroneous in a relative sense. 

Fortunately, this design flaw can be trivially remedied by augmenting the loss function with an term that captures relative error, like mean percent error:
\begin{equation}
    \mathcal{L}_{\text{ABS+\%}}(\mathbf{\theta}) = \frac{1}{N}\sum^{N}_{i=0}\left(\left\lVert \minus \nabla \hat{U}(\bm{x}_i\vert \mathbf{\theta}) - \bm{a}_i\right\rVert + \frac{\lVert \minus \nabla \hat{U}(\bm{x}_i\vert \mathbf{\theta}) - \bm{a}_i\rVert }{\lVert \bm{a}_i\rVert }\right)
    \label{eq:L_Percent}
\end{equation}
The joining of these relative and absolute error loss terms eliminates this altitude sensitivity and constitutes the second design change of the PINN-GM-III.

\subsection{Improve Numerics by Learning a Proxy to the Potential}
\label{sec:proxy_potential}

Another common problem of past machine learning gravity models is numerical clipping. These models use the same matrix operations (i.e. the neural network) to predict both very large and very small values of the potential depending on the test point's altitude, which can pose difficult numerics. When the altitude exceeds some critical threshold, the predictions grow too small, and the network prematurely clips their value to zero. This is problematic for both model training and inference --- prematurely capping the maximum altitude for which these models are usable. 

The PINN-GM-III addresses this problem by instead learning a more numerically favorable proxy to the potential, $U_{\text{NN}}$, defined as:
\begin{equation}
    U_{\text{NN}} = U * n(r); \quad n(r) = \begin{cases}
        1 \quad r < R \\
        r \quad r > R \\
    \end{cases}
    \label{eq:Potential_scaling}
\end{equation}
where $U$ is the true potential and $n(r)$ is an altitude-dependent scaling function. This scaling function leverages the fact that potentials decay according to an inverse power-law outside the Brillouin sphere. By learning a potential normalized to $1/r$, the neural network predictions will not decay to unrecoverably small values but will instead remain bounded and centered about a non-dimensionalized value of $\mu$. This scaling eliminates numerical clipping to first order, granting stable numerics out to infinity. Importantly, once the network produces its numerically stable value for $U_{\text{NN}}$, the model will then explicitly transform that value into the true potential $\hat{U}_{\text{NN}}$ by dividing the output by the known scaling function through:
\begin{equation}
    \hat{U}_{\text{NN}} = \frac{U_{\text{NN}}}{n(r)}
    \label{eq:Potential_unscaling}
\end{equation} 

\subsection{Enforcing Boundary Conditions via Transition Function to Avoid Extrapolation Error}
\label{sec:boundary_conditions}

The largest identified weakness of past machine learning gravity models is their extrapolation error. When tested beyond the bounds of their training data, machine learning models remain unconstrained and often diverge. The PINN-GM-III solves this problem by introducing a transition function that forces the model to seamlessly blend its network solution into a known analytic boundary condition outside the bounds of the training data through:
\begin{align}
    \hat{U}(r) &= w_{\text{NN}}(r)\hat{U}_{\text{NN}}(r) + w_{\text{BC}}(r)U_{\text{BC}}(r)
    \label{eq:model_blend}
\end{align}
where $\hat{U}_{\text{NN}}$ is the predicted gravitational potential, $U_{\text{BC}}$ is the potential at the known boundary condition, and $w_{\text{NN}}$ and $w_{\text{BC}}$ are the altitude dependent weights for each defined as: 
\begin{align}
    w_{\text{BC}}(r, k, r_{\text{ref}}) &= H(r, k, r_{\text{ref}}) \\
    w_{\text{NN}}(r, k, r_{\text{ref}}) &= 1 - H(r, k, r_{\text{ref}})
\end{align}
$H(r)$ is a smoothing function defined as
\begin{equation}
    H(r, k, r_{\text{ref}}) = \frac{1 + \tanh(k(r-r_{\text{ref}}))}{2}
    \label{eq:transition_fcn}
\end{equation}
where $r$ is the distance to the test point, $r_{\text{ref}}$ is a reference radius, and $k$ is a smoothing parameter which controls the sharpness of the transition. 

Equation~\eqref{eq:model_blend} takes inspiration from Reference~\citenum{zhuMachineLearningMetal2020} which introduced the idea that PINNs can enforce physics compliance through other mechanisms than terms in their loss function. Explicitly, Reference~\citenum{zhuMachineLearningMetal2020} proposed the use of smooth, differentiable forms of the Heaviside function placed at the boundary to forcibly transition neural network outputs towards the known values. 

In the case of gravity modeling, the most obvious boundary condition exists in the limit as $r \rightarrow \infty$, where the potential decays to zero. Setting $U_{\text{BC}} = 0$ and $r_{\text{ref}} = \infty$ in Equation~\eqref{eq:model_blend}, however, is not practical, as it demands that the neural network must learn a model of the potential for the entire domain $r \in [0, \infty)$. A more useful choice is to leverage insights from the spherical harmonic gravity model and recognize that high frequency components of the gravitational potential decay to zero more quickly than the point mass contribution at high altitudes --- i.e.\ 
\begin{equation}
    U_{\text{BC}}(r) = U_{\text{LF}} = \frac{\mu}{r} + \cancelto{0}{\sum_{l=0}^{n} \sum_{m=0}^{l} \frac{\mu}{r}\left(\frac{R}{r}\right)^l(\hdots)} 
\end{equation}
as $r \rightarrow \infty$. Therefore, $U_{\text{BC}}(r)$ can be set to $\frac{\mu}{r}$ assuming $r \gg R$. Therefore the PINN-GM-III sets $U_{\text{BC}} = \frac{\mu}{r} + f(r)$ in Equation~\eqref{eq:model_blend}, where $f(r)$ are any higher order terms in the spherical harmonic gravity model that the user knows a priori and wishes to leave as part of the boundary condition. 

For completeness, the recommended value for $r_{\text{ref}}$ is the maximum altitude of the training data, and the recommended value for the transition coefficient is $k=0.5$. These choices ensure that the model will quickly transition to the boundary condition outside the training data, but without applying the transition too rapidly such that it changes the gradient of the potential and accidentally induces acceleration errors. 

\subsection{Leveraging Preexisting Gravity Information into PINN-GM Solution}
\label{sec:analytic_model}
The final design change of the PINN-GM-III proposes a mechanism for incorporating past analytic models into the network solution. Explicitly, analytic models remain popular for good reason, as they are often very good at representing certain parts of the gravity field --- e.g. a point mass model captures first-order dynamics with a single parameter and and spherical harmonics can represent planetary oblateness with only $C_{2,0}$. Rather than abandoning these models and requiring the network to relearn these prominent behaviors and features, it would be far more convenient to incorporate these models into the learned solutions. 

To accomplish this, the PINN-GM-III proposes modifying Equation~\eqref{eq:model_blend} to include an analytic model term in the prediction:
\begin{equation}
    \hat{U}(r) = w_{\text{NN}}\big(U_{\text{LF}}(r) + U_{\text{NN}}(r)\big) + w_{\text{BC}}U_{\text{LF}}(r)
\end{equation} 
where $U_{\text{LF}}$ refers to the known, low-fidelity analytic model such as $U_{\text{LF}}(r) = \frac{\mu}{r} + U_{J_2}(r)$. By incorporating these low-order models, the network can exclusively focus its modeling efforts on capturing high-order perturbations from these models. 

\begin{figure}
    \centering
    \includegraphics[width=\textwidth]{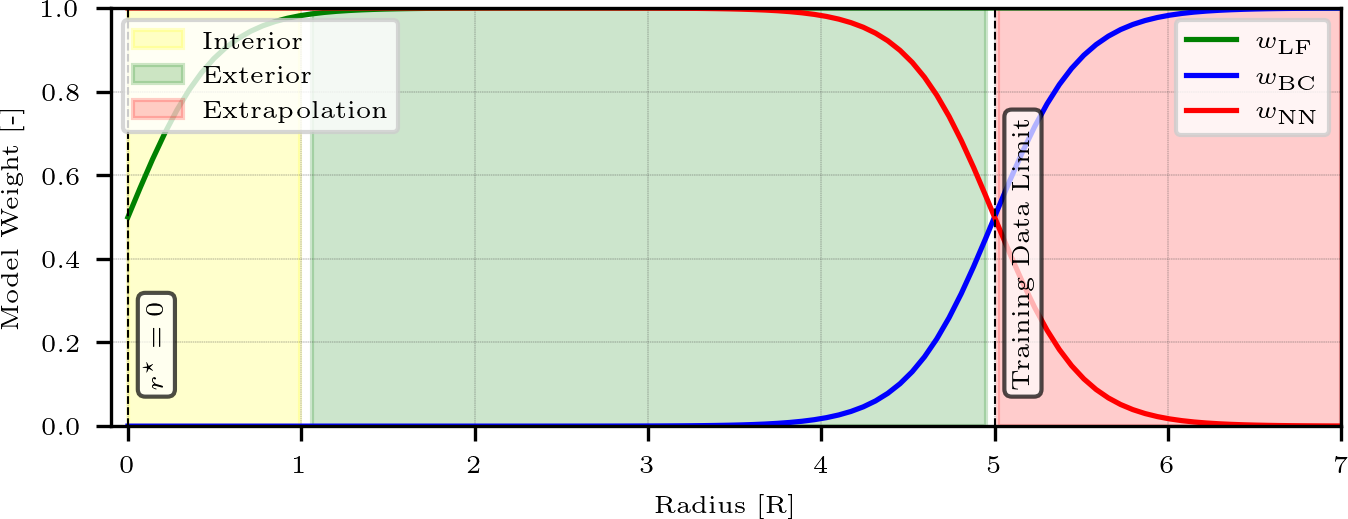}
    \caption{Visualization of the various weighting factors applied to the neural network potential, boundary condition potential, and the low-fidelity potential.}
    \label{fig:fuse_models}
\end{figure}

\begin{figure}[h!tb]
    \begin{subfigure}[b]{0.95\textwidth}
        \centering
        \includegraphics[width=\textwidth]{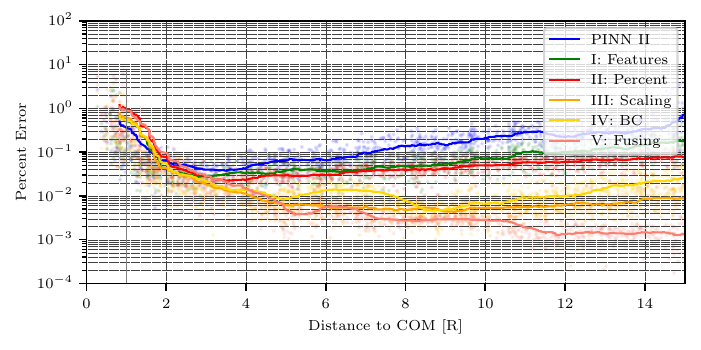}
        \caption{Error \textbf{inside} the training bounds}
        \label{fig:interpolation_mods}
    \end{subfigure}
    \begin{subfigure}[b]{0.95\textwidth}
        \centering
        \includegraphics[width=\textwidth]{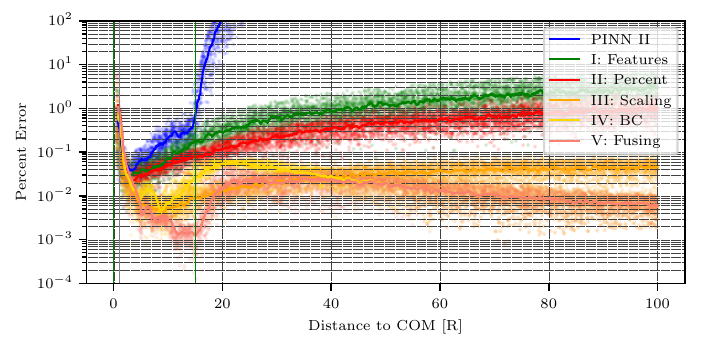}
        \caption{Error \textbf{outside} the training bounds}
        \label{fig:extrapolation_mods}
    \end{subfigure}
    \caption{Acceleration percent error as a function of altitude after sequentially applying each of the proposed PINN III modifications.}
    \label{fig:generalization_mods}
\end{figure}

It should be noted that the inclusion of analytic models should be performed carefully. In the case of small-body gravity modeling, where the geometries may vastly differ from a point mass or low-fidelity spherical harmonic approximation, the incorporation of past analytic models can induce unnecessary error at low altitudes. To ensure that these analytic models are only used where appropriate, the hyperbolic tangent fusing function, $H(x, r, k)$, is reused to dynamically weight the analytic solution through: 
\begin{equation}
    \hat{U}_{\text{LF}}(r) = w_{\text{LF}} U_{\text{LF}}(r)
\end{equation} 
where $\hat{U}_{\text{LF}}$ is the weighted low fidelity analytic model with $w_{\text{LF}}=H(r, R^*, k^*)$, $R^* = 0$, and $k=0.5$. This ensures that at high altitude, where the analytic approximation is most accurate, it fully contributes to the final learned solution, but at lower altitudes, the analytic contribution has a reduced weight to the final solution. A visualization of the various weighting functions used in the PINN-GM-III is shown in Figure~\ref{fig:fuse_models}, highlighting where the different parts of the model are fully activated and deactivated.

\subsection{Visualizing Modifications' Effects on Model Performance}
\label{sec:visualization_of_mods}

To visualize how the proposed modifications affect model accuracy, six PINN-GM are trained. The first model corresponds with the original PINN-GM-II, and the subsequent models sequentially add the proposed modifications. The PINN-GM are trained with a 5,000 point dataset uniformly distributed between the surface of the asteroid Eros to an altitude of 15R. The corresponding acceleration error of each model is reported within the training distribution in Figure~\ref{fig:interpolation_mods} and outside the training distribution in Figure~\ref{fig:extrapolation_mods}. 

Figure~\ref{fig:interpolation_mods} illustrates that these design modifications consistently improve the modeling accuracy of the PINN within the bounds of the training distribution. The PINN-GM-II had errors approaching 1\% at the high 15R altitudes; however this decreases to 0.2\%, 0.05\%, 0.01\%, 0.03\%, and 0.001\% for each of the respective modifications. Notably the transition to the new loss function (II) does reduce accuracy near the surface, but this is to be expected given the redistribution of modeling priorities that now balance accuracy at both low- and high-altitudes.

Figure~\ref{fig:extrapolation_mods} illustrates the stabilizing effect of these modifications on extrapolation error --- quantifying prediction error outside of the training bounds out to 100R. As is shown in blue, the PINN-GM-II entirely diverges after leaving the bounds of the training data at 15R; however, the design modifications stabilize performance and prevent this divergence. When only the features (I) and percent loss (II) modifications are included, the error in the high-altitude limit hits a numerical plateau as expected; however, the inclusion of the proxy potential (III) lowers that plateau by over an order-of-magnitude. The boundary condition modification (IV) eliminates this plateau --- albeit inducing a small penalty near the transition point due to changes in the gradient. Finally, the addition of the low-fidelity point mass potential (V) eliminates this penalty and further reduces the error at higher within the bounds of the training data.

\section{Benchmarking Suite}
\label{sec:case_study}

The lack of comprehensive and standard performance benchmarks for gravity models are part of the reason these failure cases had not yet been identified. To eliminate this possibility moving forward and to further scrutinize the proposed PINN-GM-III, six new evaluation metrics are proposed. These metrics characterize gravity model accuracy within and beyond the training distribution, aiming to provide both coarse and fine measures of performance in different orbital regimes. These metrics are explained in detail in Section~\ref{sec:metrics} and are used to characterize the performance of a trained PINN-GM-III on a heterogeneous density asteroid in Section~\ref{sec:case_study_experiment}. 

\subsection{Metrics}
\label{sec:metrics}

\subsubsection*{Planes Metric}
The first accuracy metric assesses the mean percent error of the predicted acceleration vector along the three cartesian planes (XY, XZ, YZ) extended between [-5R, 5R] where R is the radius of the body. The field is evaluated on a 200x200 grid of points along each plane, and the average percent error is computed as
\begin{equation}
    P = \frac{1}{N}\sum_{i=1}^N \frac{\lVert\bm{a}_{\text{true}} - \bm{a}_{\text{PINN}}\rVert}{\lVert\bm{a}_{\text{true}}\rVert} \times 100
    \label{eq:percent_error}
\end{equation}
This metric is intended to provide a coarse measure of model performance across a wide range of operational regimes.  

\subsubsection*{Generalization Metrics}
The second, third, and fourth metrics investigate the generalization of the model across a range of altitudes both within and beyond the training bounds. Explicitly, the mean acceleration error is evaluated as a function of altitude and divided into three testing regimes: interior, exterior, and extrapolation. The \textbf{interior} metric assesses error within the bounding sphere of radius R. The \textbf{exterior} metric investigates the error out to the maximum altitude of the training dataset. Finally, the \textbf{extrapolation} metric measures the error exclusively outside the training dataset --- specifically reaching altitudes 10 times larger than the maximum altitude represented in the training set. For every unit of radius, 500 samples are distributed uniformly in altitude to produce the test set. 

\subsubsection*{Surface Metric}
The fifth metric evaluates the mean acceleration error across all facets on a shape model of a celestial body, if available. This metric is used to characterize model performance at the most complex region of the field. 

\subsubsection*{Trajectory Metric}
The sixth metric evaluates the time-averaged position error of a trajectory propagated by the regressed model and the true trajectory of a spacecraft in a 24-hour low-altitude polar orbit about a rotating celestial body. Time-averaged error is used as it ensures that the error is monotonically increasing. To compute this value, the instantaneous position error $\Delta x(t)$ must be computed through 
\begin{equation}
    \Delta x(t) = \lVert \underbrace{\bm{x}(t)}_{\text{True Pos.}} - \underbrace{\bm{\hat{x}}(t)}_{\text{Propagated Pos.}} \rVert
\end{equation}
from which the time-averaged error, $\mathcal{S}$, can be computed using numerical integration via:
\begin{equation}
    \mathcal{S} = \frac{1}{T} \int_{0}^{T} \Delta x(t) \text{d}t
\end{equation}

\subsection{Experiment}
\label{sec:case_study_experiment}

These metrics are used to evaluate a PINN-GM-III trained on data generated from a synthetic, heterogeneous density asteroid modeled after 433-Eros. Heterogeneous asteroids provide an especially challenging scenario for gravity models, as their internal density distributions are not directly observable. Some asteroids contain over- and under-dense regions within their interior, or may have been formed by two asteroids merging together. While some heuristic methods have been proposed to estimate these asteroids' internal densities --- which can then be used in heterogeneous forms of the polyhedral model \cite{takahashiMorphologyDrivenDensity2014} --- the more common practice is to simply proceed with a constant density assumption. 

\begin{figure}[th!b]
    \centering
    \begin{subfigure}[b]{0.45\textwidth}
        \centering
        \includegraphics[trim=0 0 0 15, clip, width=\textwidth]{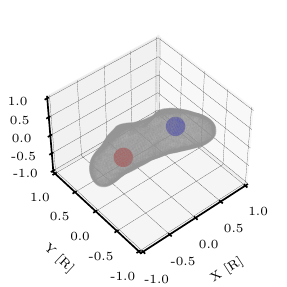}
        \caption{Heterogeneous density asteroid}    
        \label{fig:eros_gravity_field}
    \end{subfigure}
    \begin{subfigure}[b]{0.45\textwidth}
        \centering
        \includegraphics[width=\textwidth]{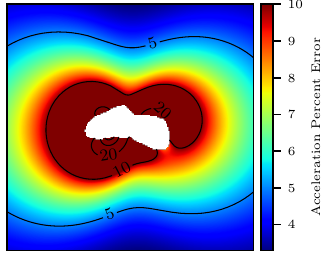}
        \caption{Constant density polyhedral error}
        \label{fig:error_planes_poly}
    \end{subfigure}
    \caption{Heterogeneous density distributions found within asteroids can cause standard models and assumptions to break down.}
\end{figure}

To induce this heterogeneous density body, two small mass inhomogeneities are placed inside the asteroid. In one hemisphere, a mass element is added, and in the other hemisphere, a mass element is removed. Each mass element contains 10\% of the total mass of the asteroid, and they are symmetrically displaced along the x-axis by 0.5R (see Figure~\ref{fig:eros_gravity_field}). The gravitational contributions of these mass elements are superimposed onto the gravity field of a constant density polyhedral model to form the simulated ground truth. This choice emulates the gravity field of a single body formed by two merged asteroids of different characteristic densities. The choice to make each mass element $\pm10$\% of the total mass is motivated based on literature with similar candidate density distributions~\cite{kanamaruEstimationInteriorDensity2019}.

Using this heterogeneous density model, 90,000 position and acceleration data are sampled uniformly around the body from 0-10R. An additional 200,700 points are sampled on the surface of the asteroid --- corresponding to a data point on every facet of the asteroid shape model. Together these approximately $300,000$ data points constitute a ``best case'' training set for the PINN-GM-III used to evaluate the upper-bound performance of the model. Subsequent sections explore model performance under more realistic and stressful data conditions. 

Using this data, a PINN-GM-II and a PINN-GM-III are trained. Both networks have six hidden layers with 32 nodes per layer which corresponds to approximately 5,300 learnable parameters. The default hyperparameters used to train these models can be found in Appendix~\ref{app:training_details} alongside the studies used to select them in Appendix~\ref{app:ablation}. Once trained, these PINN-GMs are evaluated using the aforementioned metrics and compared to a constant-density polyhedral model as shown in Figure~\ref{fig:metrics_overview}.

\subsubsection*{PINN-GM-III Performance}

\begin{figure}[ht!]
    \centering
        \begin{minipage}[l]{0.98\textwidth}
            \begin{subfigure}[b]{\textwidth}
                \centering
                \includegraphics[trim=0 5 5 0, clip, width=\textwidth]{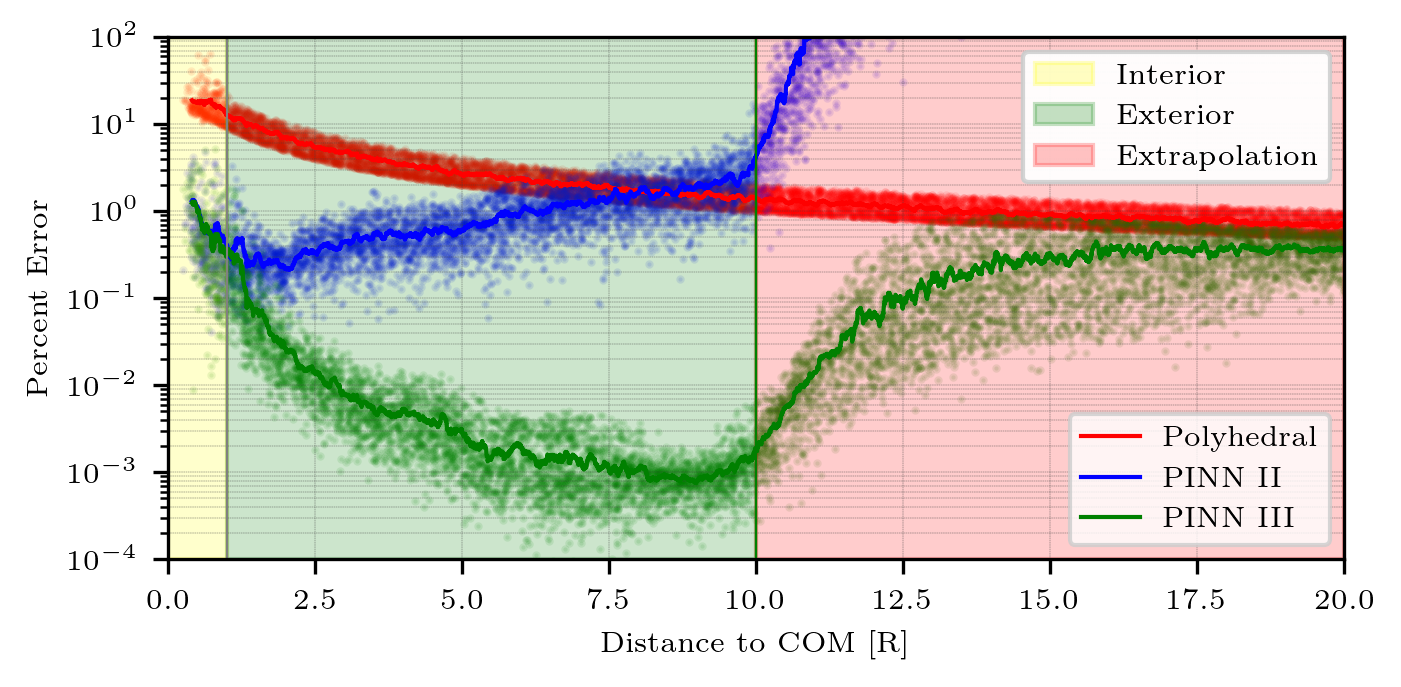}
            \end{subfigure}
        \end{minipage}
        \begin{minipage}[r]{0.01\textwidth}
            \centering
           \begin{turn}{270}
            \tiny
            \textbf{Generalization}
           \end{turn}
        \end{minipage}

        \vfill

        \begin{minipage}[l]{0.15\textwidth}
            \begin{subfigure}[b]{\textwidth}
                \includegraphics[trim=5 0 0 0, clip, width=0.95\textwidth]{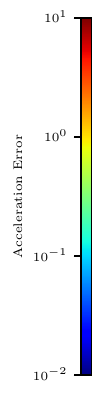}
            \end{subfigure}
        \end{minipage}
        \hfill
        \begin{minipage}[r]{0.84\textwidth}
            \begin{minipage}[l]{0.98\textwidth}
                \begin{subfigure}[b]{0.32\textwidth}
                    \centering
                    \includegraphics[width=\textwidth]{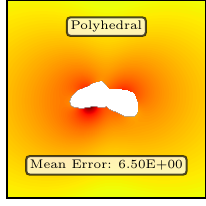}
                \end{subfigure}
                \hfill
                \begin{subfigure}[b]{0.32\textwidth}
                    \centering
                    \includegraphics[width=\textwidth]{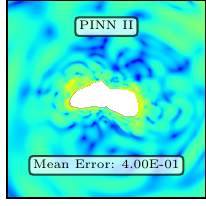}
                \end{subfigure}
                \hfill
                \begin{subfigure}[b]{0.32\textwidth}
                    \centering
                    \includegraphics[width=\textwidth]{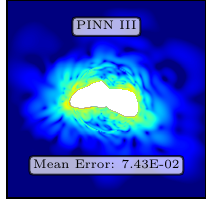}
                \end{subfigure}
            \end{minipage}
            \begin{minipage}[r]{0.01\textwidth}
                \centering
                \begin{turn}{270}
                    \tiny
                    \textbf{Planes}
                \end{turn}
            \end{minipage}

            \vfill
            \begin{minipage}[l]{0.98\textwidth}
                \begin{subfigure}[b]{0.32\textwidth}
                    \centering
                    \includegraphics[width=\textwidth]{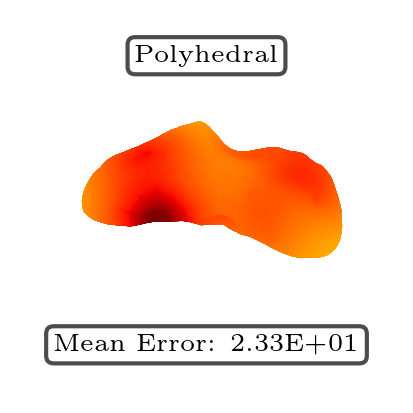}
                \end{subfigure}
                \hfill
                \begin{subfigure}[b]{0.32\textwidth}
                    \centering
                    \includegraphics[width=\textwidth]{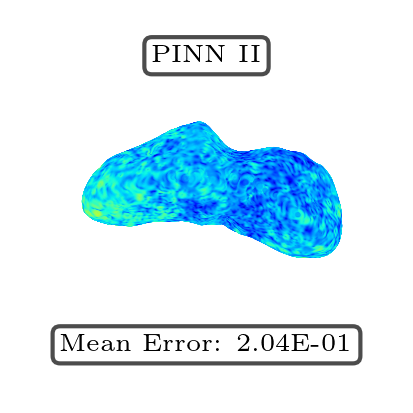}
                \end{subfigure}
                \hfill
                \begin{subfigure}[b]{0.32\textwidth}
                    \centering
                    \includegraphics[width=\textwidth]{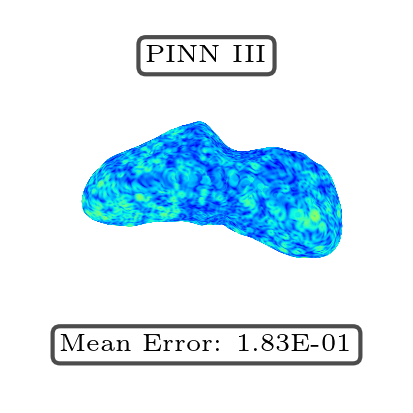}
                \end{subfigure}
            \end{minipage}
            \begin{minipage}[r]{0.01\textwidth}
                \centering
            \begin{turn}{270}
                \tiny
                \textbf{Surface}
            \end{turn}
            \end{minipage}
        \end{minipage}

        \vfill
        
        \begin{minipage}[l]{0.98\textwidth}
            \begin{subfigure}[b]{0.65\textwidth}
                \centering
                \includegraphics[width=\textwidth]{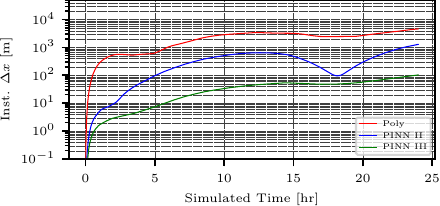}
            \end{subfigure}
            \hfill
            \begin{subfigure}[b]{0.305\textwidth}
                \centering
                \includegraphics[width=\textwidth]{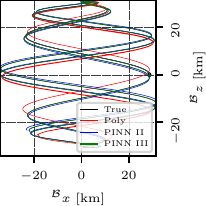}
            \end{subfigure}
        \end{minipage}
        \begin{minipage}[r]{0.01\textwidth}
                \centering
               \begin{turn}{270}
                \tiny
                \textbf{Trajectory}
               \end{turn}
        \end{minipage}
    \caption{All proposed metrics evaluated for the constant density polyhedral model, the PINN-GM-II, and the PINN-GM-III.}
    \label{fig:metrics_overview}
\end{figure}

The generalization metrics are shown at the top of Figure~\ref{fig:metrics_overview}, highlighting model performance across the three altitude regimes. The constant density polyhedral model produces the highest error on the interior and exterior metrics, averaging 10\% and 1\% respectively. In the extrapolation metric, the polyhedral error decreases as the model begins to behave like a point mass approximation. In comparison, the PINN-GM-II produces lower average errors on the interior and exterior metrics --- 0.5\% and 1\% --- however the candidate pitfalls described in Section~\ref{sec:pinn_gm} also become apparent. The former RMS loss function yields a monotonically increasing error as a function of altitude within the bounds of the training data, and the model diverges outside of it. In contrast, the PINN-GM-III maintains the lowest errors across all three regimes, averaging less than 0.5\% and 0.005\% error on the interior and exterior metrics, and maintains stability in the extrapolation regime due to the hyperbolic tangent function which enforces the point mass boundary conditions.

The planes and surface metrics are shown in the middle rows of Figure~\ref{fig:metrics_overview}. The polyhedral gravity model produced an average acceleration error of 6.5\% on the planes, the PINN-GM-II reduces this to 0.4\%, and the PINN-GM-III to 0.07\%. The errors are largest near the surface for all models, averaging at 23\% for the constant density polyhedral model, 0.2\% for the PINN-GM-II, and 0.18\% for the PINN-GM-III. 

Finally, the trajectory experiment is shown in the bottom row of Figure~\ref{fig:metrics_overview}. For this experiment, the orbit is defined by the initial conditions $\{a,e,i,\omega, \Omega, M\} = \{32\text{ km}, 0.1, 90^\circ, 0^\circ, 0^\circ, 0^\circ\}$, and the asteroid is rotating at $\omega_0=0.00073$ degrees per second along the z-axis. The figure on the left shows the instantaneous position error over the 24 hour integration period which, when time-averaged over the entire trajectory, yields average errors of 2,270 meters of error for the polyhedral model, 363 m for the PINN-GM-II, and 38 m for the PINN-GM-III. 

Under these ideal data conditions, the PINN-GM-III achieves better performance on each benchmark when compared to the constant density polyhedral model and the prior PINN-GM-II. While these results are encouraging, the robustness of these models still needs to be tested under more realistic and stressful data conditions. Moreover it remains important that the PINN-GM-III is compared to other popular gravity models beyond just previous generation PINN-GMs. To accomplish this, the next section explores what happens to these models and other models under less ideal data conditions.

\section{Comparative Study}
\label{sec:experiments}

A comparative study is performed to evaluate the PINN-GM-III against other popular gravity models. Explicitly, a point mass (PM), spherical harmonic (SH), mascon, polyhedral, extreme learning machine (ELM), traditional neural network (TNN), PINN-GM-I, PINN-GM-II, and geodesyNet model are each regressed and then evaluated using the aforementioned metrics. Each model is fit eight times, permuting the data quantity, data uncertainty, and the size of each model.\footnote{Note: The point mass model only has four permutations, as it does not have a ``large'' option and the polyhedral model only has two permutations, large and small. This is because polyhedral models are regressed in an entirely unrelated fashion to the other models --- relying primarily on image data --- and their sensitivity to data quantity and quality can not be adequately compared through this proposed experiment. Therefore, only the effect of model size is considered, and the performance can be considered an upper-bound.} These conditions purposefully test model robustness under more stressful data conditions and different model sizes.

The dataset used for the comparison is generated by sampling position and acceleration data between 0-10R about the heterogeneous density asteroid discussed in Section~\ref{sec:case_study}. For the data sparse case, $N=500$ pairs are sampled. For the data rich case, $N=50,000$ pairs are sampled. These datasets have two configurations: noisy or noiseless. The noiseless configuration assumes ideal conditions in which the acceleration vectors are perfectly observable and have no error, whereas the noisy configuration perturbs every acceleration vector in a random direction by 10\% of the acceleration magnitude via $$ \bm{\tilde{a}}_{j} = \bm{a}_j + 0.1\lVert \bm{a}_j \rVert \hat{\bm{u}_j}$$ where $\hat{\bm{u}_j}$ is randomly sampled from the unit sphere. This error is chosen as a purposefully exaggerated test case to determine which of these models remain reliable under more stressful mission conditions. 

In addition to training each model on these four datasets, this study also explores how the model size impacts performance. To do this, each model is tested at a small (S) and large (L) parametric capacity. These labels correspond to the total number of parameters used by each model---e.g. Stokes coefficients for a spherical harmonic model, facets and vertices in a shape model, or total weights and biases in a neural network. Each small model contain approximately 250 parameters, whereas the large models have approximately 30,000 parameters. Together, these varying data and parametric conditions provide insight into which models are capable of maintaining competitive performance in data sparse and low-memory regimes. Importantly, each model has slightly different regression procedures and model size calculations which are discussed in Appendix~\ref{app:regression_details}. 

\begin{table}[h!]
    \centering
    \scalebox{0.72}{
\begin{tabular}{rlllllllll}
\toprule
{} &                                      N &            \makecell{Error \\ (\%)} &                                Score &               \makecell{Planes \\ (\%)} &             \makecell{Interior \\ (\%)} &             \makecell{Exterior \\ (\%)} &              \makecell{Extrap. \\ (\%)} &               \makecell{Surface \\ (\%)} &                \makecell{Traj. \\ (km)} \\
Model               &                                        &                                     &                                      &                                         &                                         &                                         &                                         &                                          &                                         \\
\midrule
MASCONS L           &  \cellcolor[rgb]{1.00,1.00,1.00} 50000 &   \cellcolor[rgb]{1.00,1.00,1.00} 0 &   \cellcolor[rgb]{0.37,0.31,0.64} 14 &  \cellcolor[rgb]{0.37,0.31,0.64} 1.2E-2 &    \cellcolor[rgb]{0.31,0.38,0.67} 0.13 &  \cellcolor[rgb]{0.34,0.34,0.65} 5.4E-4 &  \cellcolor[rgb]{0.34,0.34,0.65} 3.4E-4 &      \cellcolor[rgb]{0.28,0.42,0.69} 8.3 &  \cellcolor[rgb]{0.35,0.34,0.65} 1.7E-3 \\
\textbf{PINN III S} &  \cellcolor[rgb]{1.00,1.00,1.00} 50000 &   \cellcolor[rgb]{1.00,1.00,1.00} 0 &   \cellcolor[rgb]{0.35,0.34,0.65} 29 &    \cellcolor[rgb]{0.29,0.41,0.69} 0.17 &     \cellcolor[rgb]{0.25,0.46,0.71} 1.2 &  \cellcolor[rgb]{0.29,0.41,0.69} 2.8E-2 &   \cellcolor[rgb]{0.31,0.39,0.67} 0.065 &       \cellcolor[rgb]{0.24,0.58,0.72} 11 &  \cellcolor[rgb]{0.27,0.44,0.70} 4.2E-2 \\
\textbf{PINN III L} &  \cellcolor[rgb]{1.00,1.00,1.00} 50000 &   \cellcolor[rgb]{1.00,1.00,1.00} 0 &   \cellcolor[rgb]{0.32,0.37,0.66} 34 &  \cellcolor[rgb]{0.34,0.34,0.65} 1.3E-2 &   \cellcolor[rgb]{0.34,0.34,0.65} 0.099 &  \cellcolor[rgb]{0.31,0.38,0.67} 1.7E-3 &     \cellcolor[rgb]{0.89,0.96,0.60} 1.3 &      \cellcolor[rgb]{0.34,0.34,0.65} 3.2 &  \cellcolor[rgb]{0.32,0.37,0.66} 2.7E-3 \\
\bottomrule
\end{tabular}

    }
    \caption{Gravity models ranked by evaluation metrics}
    \label{tab:comparison_table_score}
\end{table}

\subsubsection*{Results}

After the models are fit on the datasets, their performance is evaluated using the accuracy metrics from Section~\ref{sec:case_study} and their values are reported in Table~\ref{tab:comparison_table_score}. Along every metric (column), each model is colored by rank with the best model colored purple and the worst model colored red. Cells colored red or black correspond to values that diverged and are automatically assigned last place (70th). In the case of the planes, interior, exterior, extrapolation, and surface metrics, this divergence corresponds with values that exceed 100\% error and therefore are not usable. These individual ranks are then summed to produce a final model score which is used to sort the table and quantify relative performance.

Table~\ref{tab:comparison_table_score} should be read as follows: The first column corresponds to the specific model that was regressed, and the subsequent two columns define the data conditions used for the regression. The $N$ column is the number of training data and the error column is the magnitude of the acceleration vector error (i.e.~either 0\% or 10\%). When these cells are colored light red, that means the condition is less desirable and is used to stress model performance, whereas white corresponds to the favorable data condition. 

The fourth column is the aforementioned model score which sums the metric ranks for the listed model. For example, the small PINN-III (PINN-III S) is ranked 4th on the planes metric, 5th on the interior, 4th on exterior, 3rd on extrapolation, 8th on surface, and 5rd on the trajectory metric. Added together, these sum to a model score of 29 which is the second best score among the 70 tested models. The individual ranks for every model in each metric are reported separately in Appendix~\ref{app:regression_details} Table~\ref{tab:eros_rank_comparison_study} for reference.  

Table~\ref{tab:comparison_table_score} shows that the three highest scoring gravity models are the Mascon L, PINN-III S, and PINN-III L models respectively. Each of these models are regressed under the favorable 50,000 and 0\% error data conditions, and they all maintain low error across the different metrics. The PINN-III L performs better near the surface of the asteroid, where the mascon model is known to struggle; however, the mascon model performs better in the high-altitude extrapolation regime where no training data was present.  

Notably, five of the eight PINN-IIIs score in the top ten models, despite some of these models being trained under noisy or sparse data conditions. For example, the 5th best model is a PINN-III S, but this model is only trained with 500 data points. The next best model trained under these conditions is a Mascon L model which scores 9th and uses a two orders of magnitude more parameters. Similarly a PINN-III L performs the best among all models trained with 10\% error on the acceleration vectors --- scoring 6th across all models. Finally, a PINN-III L is also the best performer of all models trained on both sparse and noisy data, scoring in the upper 30th percentile and placing 18th overall. Table~\ref{tab:comparison_table_subset} sorts and ranks these models by their respective training conditions for a more compact summary of model performance and again highlights how PINN-III S and L remain the two highest scoring models across all tested conditions on a heterogenous density Eros.

Table~\ref{tab:comparison_table_score} simultaneously illustrates the accuracy and robustness of the PINN-GM-III to these stressful data conditions while also exposing the brittleness of other gravity models --- including past PINN-GM generations. For example, PINN-II L has the lowest error at the surface of any gravity model, averaging 2.5\% across all 200,700 facets on the shape model. However, when tested in the extrapolation regime, PINN-II diverges. Similarly, the Mascon L has the highest score across all models when regressed on ideal training conditions, but when trained on sparse and noisy data sets, this model diverges in all but one metric, ranking 58th of 70 overall. 

Past machine learning models exhibit greater concern, with the traditional neural networks, ELMs, PINN-I, and GeodesyNets consistently scoring in the lower 50th percentile. The candidate failure modes discussed in Section~\ref{sec:pinn_gm} are best captured by inspecting the interior, exterior, and extrapolation metrics. Note how these models exhibit deteriorating accuracy when evaluated on increasingly high altitude data --- i.e. the interior values are generally lower than exterior values. Moreover, these vast majority of models --- GeodesyNets excluded --- meet the divergence criteria when tested beyond the bounds of the training data. These behaviors provide confirmation of the challenges many machine models face when tested across a more diverse set of experiments, and the PINN-GM-III avoids these same challenges due to the new suite of proposed modifications. 

As a caveat, GeodesyNets are not prone to the exact same failure modes as the other machine models due to their fundamentally different architecture and evaluation procedure. Because the densities predicted are only computed within a unit volume, these models do not pose the same risk of feature or model divergence at high altitudes. That said, these models still struggle to converge as there does not exist sufficient amounts of data to reliably constraint the density function. Even with 50,000 noiseless data points, the GeodesyNets yield errors in excess of 80\% across the different testing regimes. These results further highlight how sensitive and data intensive these alternative ML model implementations can be --- illustrating the need for more nuanced discussion about how and when these models can break down. While the regression strategies used to fit these models sought to remain close to the original implementations --- as is discussed in Appendix~\ref{app:regression_details} --- it remains possible that different training configurations and practices could yield better performance. 

To ensure the conclusions drawn through this study are generalizable, three additional gravity fields are tested. These analyses regress the same 70 models on a homogeneous density version of Eros, and a homo- and a heterogenous density version of the asteroid Bennu. The imposed heterogenous density profiles again remain $\pm 10\%$ mass at locations $\pm$R$/2$ along the x-axis of the respective asteroid. These four environments attempt to span common and exotic gravity fields: from nearly spherical to irregular geometries, and from homogeneous to heterogeneous densities profiles. By studying the performance across these different cases, the general behaviors of these models can be reasonably inferred for other candidate small bodies. The exact metric values for these additional cases are reported in Appendix~\ref{app:comparison_study}, and their score summaries across the different data conditions are reported in Table~\ref{tab:table_overview_all}. 

Across these three additional experiments, the PINN-GM-III remains competitive --- consistently reporting the highest score among the machine learning models. Notably, the new scenarios do show the strength of the analytic models, such as the spherical harmonic model for the asteroid Bennu, and the polyhedral model for the constant density cases. Importantly, the results for the polyhedral model are not directly comparable to the other models reported, as the polyhedral shape was not constructed / regressed from the position and acceleration data. Its inclusion can only be used to infer the strength of the model in a parametric sense, where the polyhedral models of approximately 30,000 parameters achieves better performance than other models of the same size. Regardless, these results imply that the PINN-GM-III exhibits its greatest utility when modeling asteroids with irregular shapes and unknown density.

\begin{table}[ht!]
  \centering
      
      \begin{minipage}{0.99\textwidth}
          \begin{minipage}{0.49\textwidth}
              \raggedright
              \hspace{2.9cm}
              \includegraphics[trim=0 25 0 0, clip, width=0.40\textwidth]{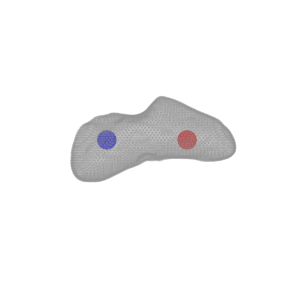}
              \vfill
              \centering
              \begin{subtable}{0.49\textwidth}
                  \centering
                  \scalebox{0.62}{\begin{tabular}{rllll}
\toprule
N & \multicolumn{2}{c}{\cellcolor[rgb]{0.99,0.79,0.71} 500} & \multicolumn{2}{c}{\cellcolor[rgb]{1.00,1.00,1.00} 50000} \\
Error (\%) &   \cellcolor[rgb]{0.99,0.79,0.71} 10 &    \cellcolor[rgb]{1.00,1.00,1.00} 0 &    \cellcolor[rgb]{0.99,0.79,0.71} 10 &   \cellcolor[rgb]{1.00,1.00,1.00} 0 \\
Model               &                                      &                                      &                                       &                                     \\
\midrule
\textbf{PINN III L} &    \cellcolor[rgb]{0.37,0.31,0.64} 1 &    \cellcolor[rgb]{0.25,0.46,0.71} 2 &     \cellcolor[rgb]{0.37,0.31,0.64} 1 &   \cellcolor[rgb]{0.23,0.57,0.72} 3 \\
\textbf{PINN III S} &    \cellcolor[rgb]{0.25,0.46,0.71} 2 &    \cellcolor[rgb]{0.37,0.31,0.64} 1 &     \cellcolor[rgb]{0.75,0.90,0.63} 6 &   \cellcolor[rgb]{0.27,0.44,0.70} 2 \\
PINN II L           &    \cellcolor[rgb]{0.26,0.61,0.71} 3 &    \cellcolor[rgb]{0.40,0.76,0.65} 4 &     \cellcolor[rgb]{0.26,0.61,0.71} 3 &   \cellcolor[rgb]{0.35,0.71,0.67} 4 \\
MASCONS L           &   \cellcolor[rgb]{0.76,0.16,0.29} 15 &    \cellcolor[rgb]{0.26,0.61,0.71} 3 &     \cellcolor[rgb]{0.25,0.46,0.71} 2 &   \cellcolor[rgb]{0.37,0.31,0.64} 1 \\
PINN II S           &    \cellcolor[rgb]{0.40,0.76,0.65} 4 &    \cellcolor[rgb]{0.75,0.90,0.63} 6 &     \cellcolor[rgb]{0.58,0.83,0.64} 5 &   \cellcolor[rgb]{0.50,0.80,0.65} 5 \\
MASCONS S           &    \cellcolor[rgb]{1.00,0.96,0.68} 9 &    \cellcolor[rgb]{0.58,0.83,0.64} 5 &     \cellcolor[rgb]{0.40,0.76,0.65} 4 &   \cellcolor[rgb]{0.79,0.92,0.62} 7 \\
POLY S              &  \cellcolor[rgb]{0.00,0.00,0.00} inf &  \cellcolor[rgb]{0.00,0.00,0.00} inf &   \cellcolor[rgb]{0.00,0.00,0.00} inf &   \cellcolor[rgb]{0.65,0.86,0.64} 6 \\
PINN I L            &    \cellcolor[rgb]{0.75,0.90,0.63} 6 &   \cellcolor[rgb]{0.96,0.43,0.26} 13 &     \cellcolor[rgb]{0.90,0.96,0.60} 7 &   \cellcolor[rgb]{0.97,0.99,0.70} 9 \\
PM -                &    \cellcolor[rgb]{0.58,0.83,0.64} 5 &    \cellcolor[rgb]{0.90,0.96,0.60} 7 &    \cellcolor[rgb]{1.00,0.88,0.55} 10 &  \cellcolor[rgb]{0.99,0.79,0.47} 12 \\
POLY L              &  \cellcolor[rgb]{0.00,0.00,0.00} inf &  \cellcolor[rgb]{0.00,0.00,0.00} inf &   \cellcolor[rgb]{0.00,0.00,0.00} inf &   \cellcolor[rgb]{0.91,0.97,0.61} 8 \\
SH S                &    \cellcolor[rgb]{0.90,0.96,0.60} 7 &    \cellcolor[rgb]{0.97,0.99,0.70} 8 &     \cellcolor[rgb]{1.00,0.96,0.68} 9 &  \cellcolor[rgb]{1.00,0.96,0.69} 10 \\
PINN I S            &    \cellcolor[rgb]{0.97,0.99,0.70} 8 &    \cellcolor[rgb]{1.00,0.96,0.68} 9 &     \cellcolor[rgb]{0.97,0.99,0.70} 8 &  \cellcolor[rgb]{1.00,0.89,0.57} 11 \\
GEONET L            &   \cellcolor[rgb]{1.00,0.88,0.55} 10 &   \cellcolor[rgb]{1.00,0.88,0.55} 10 &    \cellcolor[rgb]{0.98,0.60,0.34} 12 &  \cellcolor[rgb]{0.97,0.52,0.30} 14 \\
SH L                &   \cellcolor[rgb]{0.99,0.75,0.44} 11 &   \cellcolor[rgb]{0.98,0.60,0.34} 12 &    \cellcolor[rgb]{0.99,0.75,0.44} 11 &  \cellcolor[rgb]{0.99,0.67,0.37} 13 \\
GEONET S            &   \cellcolor[rgb]{0.98,0.60,0.34} 12 &   \cellcolor[rgb]{0.99,0.75,0.44} 11 &    \cellcolor[rgb]{0.96,0.43,0.26} 13 &  \cellcolor[rgb]{0.93,0.38,0.27} 15 \\
TNN S               &   \cellcolor[rgb]{0.62,0.00,0.26} 17 &   \cellcolor[rgb]{0.62,0.00,0.26} 16 &    \cellcolor[rgb]{0.88,0.30,0.29} 14 &  \cellcolor[rgb]{0.86,0.28,0.30} 16 \\
ELM L               &   \cellcolor[rgb]{0.96,0.43,0.26} 13 &   \cellcolor[rgb]{0.88,0.30,0.29} 14 &    \cellcolor[rgb]{0.62,0.00,0.26} 16 &  \cellcolor[rgb]{0.62,0.00,0.26} 18 \\
ELM S               &   \cellcolor[rgb]{0.88,0.30,0.29} 14 &   \cellcolor[rgb]{0.76,0.16,0.29} 15 &    \cellcolor[rgb]{0.62,0.00,0.26} 17 &  \cellcolor[rgb]{0.62,0.00,0.26} 19 \\
TNN L               &   \cellcolor[rgb]{0.62,0.00,0.26} 16 &   \cellcolor[rgb]{0.62,0.00,0.26} 17 &    \cellcolor[rgb]{0.76,0.16,0.29} 15 &  \cellcolor[rgb]{0.75,0.14,0.29} 17 \\
\bottomrule
\end{tabular}
}
                  \captionsetup{font=footnotesize}
                  \caption{\scalebox{0.9}{Eros Heterogeneous}}
                  \label{tab:comparison_table_subset}
              \end{subtable}
          \end{minipage}
          \begin{minipage}{0.49\textwidth}
              \raggedright
              \hspace{2.9cm}
              \includegraphics[trim=0 25 0 0, clip, width=0.40\textwidth]{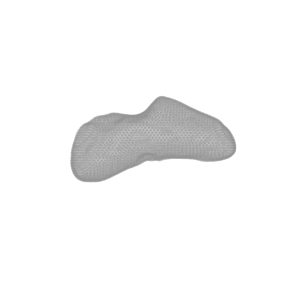}
              \vfill
              \centering
              \begin{subtable}{0.49\textwidth}
                  \centering
                  \scalebox{0.62}{\begin{tabular}{rllll}
\toprule
N & \multicolumn{2}{c}{\cellcolor[rgb]{0.99,0.79,0.71} 500} & \multicolumn{2}{c}{\cellcolor[rgb]{1.00,1.00,1.00} 50000} \\
Error (\%) &   \cellcolor[rgb]{0.99,0.79,0.71} 10 &    \cellcolor[rgb]{1.00,1.00,1.00} 0 &    \cellcolor[rgb]{0.99,0.79,0.71} 10 &   \cellcolor[rgb]{1.00,1.00,1.00} 0 \\
Model               &                                      &                                      &                                       &                                     \\
\midrule
POLY L              &  \cellcolor[rgb]{0.00,0.00,0.00} inf &  \cellcolor[rgb]{0.00,0.00,0.00} inf &   \cellcolor[rgb]{0.00,0.00,0.00} inf &   \cellcolor[rgb]{0.37,0.31,0.64} 1 \\
\textbf{PINN III S} &    \cellcolor[rgb]{0.37,0.31,0.64} 2 &    \cellcolor[rgb]{0.37,0.31,0.64} 2 &     \cellcolor[rgb]{0.37,0.31,0.64} 1 &   \cellcolor[rgb]{0.35,0.71,0.67} 4 \\
\textbf{PINN III L} &    \cellcolor[rgb]{0.37,0.31,0.64} 2 &    \cellcolor[rgb]{0.37,0.31,0.64} 2 &     \cellcolor[rgb]{0.25,0.46,0.71} 2 &   \cellcolor[rgb]{0.23,0.57,0.72} 3 \\
PINN II L           &    \cellcolor[rgb]{0.25,0.47,0.71} 3 &    \cellcolor[rgb]{0.28,0.63,0.70} 4 &     \cellcolor[rgb]{0.58,0.83,0.64} 5 &   \cellcolor[rgb]{0.65,0.86,0.64} 6 \\
POLY S              &  \cellcolor[rgb]{0.00,0.00,0.00} inf &  \cellcolor[rgb]{0.00,0.00,0.00} inf &   \cellcolor[rgb]{0.00,0.00,0.00} inf &   \cellcolor[rgb]{0.50,0.80,0.65} 5 \\
PINN II S           &    \cellcolor[rgb]{0.28,0.63,0.70} 4 &    \cellcolor[rgb]{0.63,0.85,0.64} 6 &     \cellcolor[rgb]{0.75,0.90,0.63} 6 &   \cellcolor[rgb]{0.79,0.92,0.62} 7 \\
MASCONS L           &   \cellcolor[rgb]{0.77,0.17,0.29} 15 &    \cellcolor[rgb]{0.25,0.47,0.71} 3 &     \cellcolor[rgb]{0.26,0.61,0.71} 3 &   \cellcolor[rgb]{0.27,0.44,0.70} 2 \\
MASCONS S           &    \cellcolor[rgb]{0.93,0.97,0.64} 8 &    \cellcolor[rgb]{0.43,0.77,0.65} 5 &     \cellcolor[rgb]{0.40,0.76,0.65} 4 &   \cellcolor[rgb]{0.91,0.97,0.61} 8 \\
PM -                &    \cellcolor[rgb]{0.43,0.77,0.65} 5 &    \cellcolor[rgb]{0.80,0.92,0.62} 7 &     \cellcolor[rgb]{1.00,0.96,0.68} 9 &  \cellcolor[rgb]{0.99,0.79,0.47} 12 \\
PINN I L            &    \cellcolor[rgb]{0.80,0.92,0.62} 7 &   \cellcolor[rgb]{0.96,0.46,0.28} 13 &     \cellcolor[rgb]{0.90,0.96,0.60} 7 &   \cellcolor[rgb]{0.97,0.99,0.70} 9 \\
PINN I S            &    \cellcolor[rgb]{1.00,1.00,0.75} 9 &   \cellcolor[rgb]{0.99,0.79,0.47} 11 &     \cellcolor[rgb]{0.97,0.99,0.70} 8 &  \cellcolor[rgb]{1.00,0.96,0.69} 10 \\
SH S                &    \cellcolor[rgb]{0.80,0.92,0.62} 7 &    \cellcolor[rgb]{0.93,0.97,0.64} 8 &    \cellcolor[rgb]{1.00,0.88,0.55} 10 &  \cellcolor[rgb]{1.00,0.89,0.57} 11 \\
GEONET L            &   \cellcolor[rgb]{1.00,0.91,0.60} 10 &    \cellcolor[rgb]{1.00,1.00,0.75} 9 &    \cellcolor[rgb]{0.99,0.75,0.44} 11 &  \cellcolor[rgb]{0.99,0.67,0.37} 13 \\
GEONET S            &   \cellcolor[rgb]{0.99,0.65,0.36} 12 &   \cellcolor[rgb]{0.99,0.79,0.47} 11 &    \cellcolor[rgb]{0.96,0.43,0.26} 13 &  \cellcolor[rgb]{0.93,0.38,0.27} 15 \\
SH L                &   \cellcolor[rgb]{0.99,0.79,0.47} 11 &   \cellcolor[rgb]{0.99,0.65,0.36} 12 &    \cellcolor[rgb]{0.98,0.60,0.34} 12 &  \cellcolor[rgb]{0.97,0.52,0.30} 14 \\
TNN S               &   \cellcolor[rgb]{0.62,0.00,0.26} 16 &   \cellcolor[rgb]{0.62,0.00,0.26} 16 &    \cellcolor[rgb]{0.88,0.30,0.29} 14 &  \cellcolor[rgb]{0.75,0.14,0.29} 17 \\
ELM L               &   \cellcolor[rgb]{0.96,0.46,0.28} 13 &   \cellcolor[rgb]{0.89,0.32,0.29} 14 &    \cellcolor[rgb]{0.62,0.00,0.26} 16 &  \cellcolor[rgb]{0.62,0.00,0.26} 18 \\
ELM S               &   \cellcolor[rgb]{0.89,0.32,0.29} 14 &   \cellcolor[rgb]{0.77,0.17,0.29} 15 &    \cellcolor[rgb]{0.62,0.00,0.26} 17 &  \cellcolor[rgb]{0.62,0.00,0.26} 19 \\
TNN L               &   \cellcolor[rgb]{0.62,0.00,0.26} 17 &   \cellcolor[rgb]{0.62,0.00,0.26} 17 &    \cellcolor[rgb]{0.76,0.16,0.29} 15 &  \cellcolor[rgb]{0.86,0.28,0.30} 16 \\
\bottomrule
\end{tabular}
}
                  \captionsetup{font=footnotesize}
                  \caption{\scalebox{0.9}{Eros Homogeneous}}
              \end{subtable}
          \end{minipage}
      \end{minipage}
      \vfill
      \begin{minipage}{0.98\textwidth}
          \begin{minipage}{0.49\textwidth}
              \raggedright
              \hspace{2.9cm}
              \includegraphics[trim=0 25 0 0, clip, width=0.40\textwidth]{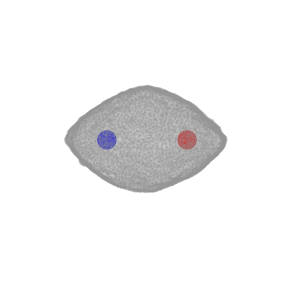}
              \vfill
              \centering
              \begin{subtable}{0.49\textwidth}
                  \centering
                  \scalebox{0.62}{\begin{tabular}{rllll}
\toprule
N & \multicolumn{2}{c}{\cellcolor[rgb]{0.99,0.79,0.71} 500} & \multicolumn{2}{c}{\cellcolor[rgb]{1.00,1.00,1.00} 50000} \\
Error (\%) &   \cellcolor[rgb]{0.99,0.79,0.71} 10 &    \cellcolor[rgb]{1.00,1.00,1.00} 0 &    \cellcolor[rgb]{0.99,0.79,0.71} 10 &   \cellcolor[rgb]{1.00,1.00,1.00} 0 \\
Model               &                                      &                                      &                                       &                                     \\
\midrule
SH S                &    \cellcolor[rgb]{0.37,0.31,0.64} 1 &    \cellcolor[rgb]{0.37,0.31,0.64} 1 &     \cellcolor[rgb]{0.37,0.31,0.64} 1 &   \cellcolor[rgb]{0.35,0.71,0.67} 4 \\
\textbf{PINN III L} &    \cellcolor[rgb]{0.25,0.46,0.71} 2 &    \cellcolor[rgb]{0.25,0.46,0.71} 2 &     \cellcolor[rgb]{0.40,0.76,0.65} 4 &   \cellcolor[rgb]{0.27,0.44,0.70} 2 \\
\textbf{PINN III S} &    \cellcolor[rgb]{0.58,0.83,0.64} 5 &    \cellcolor[rgb]{0.26,0.61,0.71} 3 &     \cellcolor[rgb]{0.90,0.96,0.60} 7 &   \cellcolor[rgb]{0.23,0.57,0.72} 3 \\
PINN II L           &    \cellcolor[rgb]{0.26,0.61,0.71} 3 &    \cellcolor[rgb]{0.40,0.76,0.65} 4 &     \cellcolor[rgb]{0.75,0.90,0.63} 6 &   \cellcolor[rgb]{0.50,0.80,0.65} 5 \\
PINN II S           &    \cellcolor[rgb]{0.75,0.90,0.63} 6 &    \cellcolor[rgb]{0.75,0.90,0.63} 6 &     \cellcolor[rgb]{0.26,0.61,0.71} 3 &   \cellcolor[rgb]{0.65,0.86,0.64} 6 \\
MASCONS L           &   \cellcolor[rgb]{0.76,0.16,0.29} 15 &    \cellcolor[rgb]{0.58,0.83,0.64} 5 &     \cellcolor[rgb]{0.58,0.83,0.64} 5 &   \cellcolor[rgb]{0.37,0.31,0.64} 1 \\
PM -                &    \cellcolor[rgb]{0.40,0.76,0.65} 4 &    \cellcolor[rgb]{0.90,0.96,0.60} 7 &     \cellcolor[rgb]{1.00,0.96,0.68} 9 &  \cellcolor[rgb]{1.00,0.89,0.57} 11 \\
MASCONS S           &   \cellcolor[rgb]{0.88,0.30,0.29} 14 &   \cellcolor[rgb]{1.00,0.88,0.55} 10 &     \cellcolor[rgb]{0.25,0.46,0.71} 2 &   \cellcolor[rgb]{0.79,0.92,0.62} 7 \\
PINN I L            &    \cellcolor[rgb]{0.90,0.96,0.60} 7 &    \cellcolor[rgb]{0.97,0.99,0.70} 8 &     \cellcolor[rgb]{0.97,0.99,0.70} 8 &  \cellcolor[rgb]{1.00,0.96,0.69} 10 \\
POLY L              &  \cellcolor[rgb]{0.00,0.00,0.00} inf &  \cellcolor[rgb]{0.00,0.00,0.00} inf &   \cellcolor[rgb]{0.00,0.00,0.00} inf &   \cellcolor[rgb]{0.91,0.97,0.61} 8 \\
PINN I S            &    \cellcolor[rgb]{1.00,0.96,0.68} 9 &   \cellcolor[rgb]{0.98,0.60,0.34} 12 &    \cellcolor[rgb]{1.00,0.88,0.55} 10 &  \cellcolor[rgb]{0.99,0.79,0.47} 12 \\
POLY S              &  \cellcolor[rgb]{0.00,0.00,0.00} inf &  \cellcolor[rgb]{0.00,0.00,0.00} inf &   \cellcolor[rgb]{0.00,0.00,0.00} inf &  \cellcolor[rgb]{1.00,0.96,0.69} 10 \\
SH L                &    \cellcolor[rgb]{0.97,0.99,0.70} 8 &    \cellcolor[rgb]{1.00,0.96,0.68} 9 &    \cellcolor[rgb]{0.99,0.75,0.44} 11 &  \cellcolor[rgb]{0.99,0.67,0.37} 13 \\
GEONET L            &   \cellcolor[rgb]{1.00,0.88,0.55} 10 &   \cellcolor[rgb]{0.99,0.75,0.44} 11 &    \cellcolor[rgb]{0.98,0.60,0.34} 12 &  \cellcolor[rgb]{0.97,0.52,0.30} 14 \\
GEONET S            &   \cellcolor[rgb]{0.99,0.75,0.44} 11 &   \cellcolor[rgb]{0.96,0.43,0.26} 13 &    \cellcolor[rgb]{0.96,0.43,0.26} 13 &  \cellcolor[rgb]{0.93,0.38,0.27} 15 \\
TNN L               &   \cellcolor[rgb]{0.62,0.00,0.26} 16 &   \cellcolor[rgb]{0.62,0.00,0.26} 17 &    \cellcolor[rgb]{0.88,0.30,0.29} 14 &  \cellcolor[rgb]{0.86,0.28,0.30} 16 \\
ELM L               &   \cellcolor[rgb]{0.98,0.60,0.34} 12 &   \cellcolor[rgb]{0.76,0.16,0.29} 15 &    \cellcolor[rgb]{0.62,0.00,0.26} 16 &  \cellcolor[rgb]{0.62,0.00,0.26} 18 \\
ELM S               &   \cellcolor[rgb]{0.88,0.30,0.29} 14 &   \cellcolor[rgb]{0.88,0.30,0.29} 14 &    \cellcolor[rgb]{0.62,0.00,0.26} 17 &  \cellcolor[rgb]{0.62,0.00,0.26} 19 \\
TNN S               &   \cellcolor[rgb]{0.62,0.00,0.26} 17 &   \cellcolor[rgb]{0.62,0.00,0.26} 16 &    \cellcolor[rgb]{0.76,0.16,0.29} 15 &  \cellcolor[rgb]{0.75,0.14,0.29} 17 \\
\bottomrule
\end{tabular}
}
                  \captionsetup{font=footnotesize}
                  \caption{\scalebox{0.9}{Bennu Heterogeneous}}
              \end{subtable}
          \end{minipage}
          \begin{minipage}{0.49\textwidth}
              \begin{minipage}{\textwidth}
                  \raggedright
                  \hspace{2.9cm}
                  \includegraphics[trim=0 25 0 0, clip, width=0.40\textwidth]{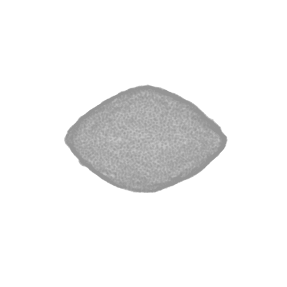}
              \end{minipage}
              \vfill
              \centering
              \begin{subtable}{0.49\textwidth}
                  \centering
                  \scalebox{0.62}{\begin{tabular}{rllll}
\toprule
N & \multicolumn{2}{c}{\cellcolor[rgb]{0.99,0.79,0.71} 500} & \multicolumn{2}{c}{\cellcolor[rgb]{1.00,1.00,1.00} 50000} \\
Error (\%) &   \cellcolor[rgb]{0.99,0.79,0.71} 10 &    \cellcolor[rgb]{1.00,1.00,1.00} 0 &    \cellcolor[rgb]{0.99,0.79,0.71} 10 &   \cellcolor[rgb]{1.00,1.00,1.00} 0 \\
Model               &                                      &                                      &                                       &                                     \\
\midrule
POLY L              &  \cellcolor[rgb]{0.00,0.00,0.00} inf &  \cellcolor[rgb]{0.00,0.00,0.00} inf &   \cellcolor[rgb]{0.00,0.00,0.00} inf &   \cellcolor[rgb]{0.37,0.31,0.64} 1 \\
SH S                &    \cellcolor[rgb]{0.37,0.31,0.64} 1 &    \cellcolor[rgb]{0.37,0.31,0.64} 1 &     \cellcolor[rgb]{0.37,0.31,0.64} 1 &   \cellcolor[rgb]{0.65,0.86,0.64} 6 \\
\textbf{PINN III S} &    \cellcolor[rgb]{0.40,0.76,0.65} 4 &    \cellcolor[rgb]{0.40,0.76,0.65} 4 &     \cellcolor[rgb]{0.25,0.46,0.71} 2 &   \cellcolor[rgb]{0.35,0.71,0.67} 4 \\
\textbf{PINN III L} &    \cellcolor[rgb]{0.40,0.76,0.65} 4 &    \cellcolor[rgb]{0.25,0.46,0.71} 2 &     \cellcolor[rgb]{0.40,0.76,0.65} 4 &   \cellcolor[rgb]{0.35,0.71,0.67} 4 \\
POLY S              &  \cellcolor[rgb]{0.00,0.00,0.00} inf &  \cellcolor[rgb]{0.00,0.00,0.00} inf &   \cellcolor[rgb]{0.00,0.00,0.00} inf &   \cellcolor[rgb]{0.50,0.80,0.65} 5 \\
PINN II L           &    \cellcolor[rgb]{0.58,0.83,0.64} 5 &    \cellcolor[rgb]{0.58,0.83,0.64} 5 &     \cellcolor[rgb]{0.75,0.90,0.63} 6 &   \cellcolor[rgb]{0.79,0.92,0.62} 7 \\
PINN II S           &    \cellcolor[rgb]{0.75,0.90,0.63} 6 &    \cellcolor[rgb]{0.75,0.90,0.63} 6 &     \cellcolor[rgb]{0.58,0.83,0.64} 5 &   \cellcolor[rgb]{0.91,0.97,0.61} 8 \\
MASCONS L           &   \cellcolor[rgb]{0.76,0.16,0.29} 15 &    \cellcolor[rgb]{0.26,0.61,0.71} 3 &     \cellcolor[rgb]{0.90,0.96,0.60} 7 &   \cellcolor[rgb]{0.27,0.44,0.70} 2 \\
PM -                &    \cellcolor[rgb]{0.25,0.46,0.71} 2 &    \cellcolor[rgb]{0.90,0.96,0.60} 7 &     \cellcolor[rgb]{0.97,0.99,0.70} 8 &  \cellcolor[rgb]{1.00,0.89,0.57} 11 \\
PINN I L            &    \cellcolor[rgb]{0.97,0.99,0.70} 8 &    \cellcolor[rgb]{0.97,0.99,0.70} 8 &     \cellcolor[rgb]{1.00,0.96,0.68} 9 &   \cellcolor[rgb]{0.97,0.99,0.70} 9 \\
MASCONS S           &   \cellcolor[rgb]{0.96,0.43,0.26} 13 &   \cellcolor[rgb]{1.00,0.88,0.55} 10 &     \cellcolor[rgb]{0.26,0.61,0.71} 3 &  \cellcolor[rgb]{1.00,0.96,0.69} 10 \\
PINN I S            &    \cellcolor[rgb]{1.00,0.96,0.68} 9 &   \cellcolor[rgb]{0.98,0.60,0.34} 12 &    \cellcolor[rgb]{1.00,0.88,0.55} 10 &  \cellcolor[rgb]{0.99,0.79,0.47} 12 \\
SH L                &    \cellcolor[rgb]{0.90,0.96,0.60} 7 &    \cellcolor[rgb]{1.00,0.96,0.68} 9 &    \cellcolor[rgb]{0.98,0.60,0.34} 12 &  \cellcolor[rgb]{0.99,0.67,0.37} 13 \\
GEONET L            &   \cellcolor[rgb]{1.00,0.88,0.55} 10 &   \cellcolor[rgb]{0.99,0.75,0.44} 11 &    \cellcolor[rgb]{0.99,0.75,0.44} 11 &  \cellcolor[rgb]{0.97,0.52,0.30} 14 \\
GEONET S            &   \cellcolor[rgb]{0.99,0.75,0.44} 11 &   \cellcolor[rgb]{0.96,0.43,0.26} 13 &    \cellcolor[rgb]{0.96,0.43,0.26} 13 &  \cellcolor[rgb]{0.93,0.38,0.27} 15 \\
ELM L               &   \cellcolor[rgb]{0.98,0.60,0.34} 12 &   \cellcolor[rgb]{0.88,0.30,0.29} 14 &    \cellcolor[rgb]{0.62,0.00,0.26} 16 &  \cellcolor[rgb]{0.62,0.00,0.26} 18 \\
ELM S               &   \cellcolor[rgb]{0.88,0.30,0.29} 14 &   \cellcolor[rgb]{0.76,0.16,0.29} 15 &    \cellcolor[rgb]{0.62,0.00,0.26} 17 &  \cellcolor[rgb]{0.62,0.00,0.26} 19 \\
TNN L               &   \cellcolor[rgb]{0.62,0.00,0.26} 17 &   \cellcolor[rgb]{0.62,0.00,0.26} 16 &    \cellcolor[rgb]{0.88,0.30,0.29} 14 &  \cellcolor[rgb]{0.75,0.14,0.29} 17 \\
TNN S               &   \cellcolor[rgb]{0.62,0.00,0.26} 16 &   \cellcolor[rgb]{0.62,0.00,0.26} 17 &    \cellcolor[rgb]{0.76,0.16,0.29} 15 &  \cellcolor[rgb]{0.86,0.28,0.30} 16 \\
\bottomrule
\end{tabular}
}
                  \captionsetup{font=footnotesize}
                  \caption{\scalebox{0.9}{Bennu Homogeneous}}
              \end{subtable}
          \end{minipage}
      \end{minipage}

  \caption{Summarized rank for all models across four test asteroids}
  \label{tab:table_overview_all}
\end{table}

\subsubsection*{Inference Time}

\begin{figure}[!htbp]
 \centering
  \includegraphics[width=\textwidth]{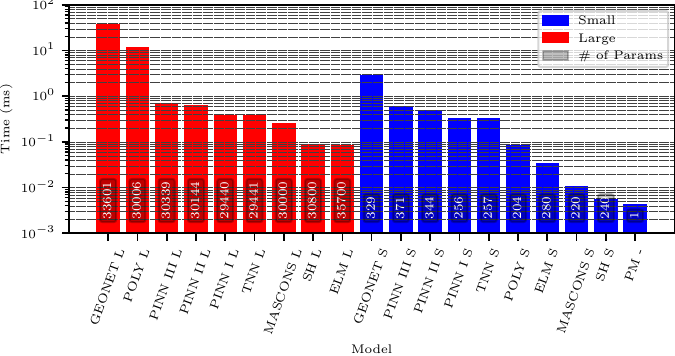}
  \caption{Inference time per sample with the unique number of parameters in each model overlaid on the individual bars. }
  \label{fig:inference_time}
\end{figure}

While Tables~\ref{tab:comparison_table_score} and \ref{tab:comparison_table_subset} highlight the relative accuracy of the available gravity models, another important comparison point is each model's evaluation speed. To characterize this, 1,000 randomly distributed test points are evaluated using each gravity model. The total evaluation time is measured, and an average inference time per sample is reported in Figure~\ref{fig:inference_time}.

For the large models, Figure~\ref{fig:inference_time} shows that the large GeodesyNet (33,601 parameters) and the large polyhedral model (30,006 parameters / approximately 10,000 facets) are the slowest gravity models reporting approximately 40 and 10 ms for each evaluation respectively. The neural network models (PINN I-III and TNN) execute an order of magnitude faster, executing between 0.4 - 0.7 ms, followed by the mascon, spherical harmonic, and ELM models at 0.3 ms, 0.09 ms, and 0.08 ms respectively. For small models, the GeodesyNet is the slowest at 3 ms, followed by the remaining machine learning gravity models averaging between 0.3 and 0.6 ms. The small polyhedral model (204 parameters or 66 faces) is faster at 0.09 ms, and the mascon, spherical harmonic, and point mass models are the fastest, executing in less than 0.01 ms per evaluation. 

These results highlight that speed is among the largest trade-offs for the PINN-GM-III, particularly in the small model regime. That said, the PINN-III models are not prohibitively slow, and their computational cost remains nearly constant irrespective of model size. For example, PINN-III L is two orders of magnitude larger than PINN-III S, but its computational cost is only twice as large. In contrast, the large polyhedral models cost appears to scale linearly, taking approximately 100 times longer for a model that is 100 times bigger. This suggests that the PINN-III remains a strong option if robustness and accuracy are the primary goal, and users can increase the size and performance of these models with relatively little added overhead. 

\section{Conclusions}
\label{sec:conclusions}

Scientific machine learning and physics informed neural networks offer a compelling set of tools to address the gravity field modeling problem. Rather than using prescriptive analytic gravity models prone to various limitations, PINNs can learn convenient representations of the gravitational potential while maintaining desirable physics properties and assurances. While past machine learning gravity models have offered early glimpses into the potential advantages of these numerical solutions, this greater class of model are often susceptible to various pitfalls that had yet to be exposed or addressed. This paper highlights these failure cases for a variety of past models, and introduces design modifications within the machine learning architecture that can overcome these challenges. Taken together, these modifications form the third generation PINN gravity model, or PINN-GM-III. This model is designed to solve the problems of feature divergence, bias towards low-altitude samples, numerical instability, and extrapolation error, while also proposing a framework for fusing analytic and numerical gravity models together for enhanced modeling accuracy. While these modifications are studied exclusively on the PINN gravity model, it should be noted that many of these modifications can likely be applied to other machine learning solutions to enhance their own robustness and generalizability.

Beyond introducing the PINN-GM-III, this manuscript also proposes new evaluation metrics to more comprehensively assess the modeling accuracy of various analytic and numerical gravity models in different orbital regimes. When trained on data from both homogeneous and heterogeneous density asteroids, the PINN-GM-III is shown to achieve competitive performance over past generations and other numerical and analytic models, while also demonstrating robustness to data sparse and noisy conditions. Future work will continue to investigate design modifications that can improve model performance, with a particular emphasis on returning to large celestial bodies and investigating ways in which the high-frequency components can be learned and represented more efficiently.

\section{Statements and Declarations}
On behalf of all authors, the corresponding author states that there is no conflict of interest. 

\section{Acknowledgements}
This work utilized the Alpine high performance computing resource at the University of Colorado Boulder. Alpine is jointly funded by the University of Colorado Boulder, the University of Colorado Anschutz, Colorado State University, and the National Science Foundation (award 2201538).

\bibliography{bibliography.bib}

\appendix

\section{PINN-GM-III Training Details}
\label{app:training_details}
Like the PINN-GM-II, the PINN-GM-III is composed of a feed-forward multi-layer perception, preceded by a feature engineering / embedding layer. Skip connections are attached between the embedding layer and each of the hidden layers of the network, and the final layer uses linear activation functions to produce the network's prediction of the proxy potential. Unlike the PINN-GM-II, all of the experiments tested in this paper do not make use of the multi-constraint loss function due to findings presented in Appendix~\ref{app:ablation}.

The default hyperparameters used to train PINN-GM-III are listed in Table~\ref{tab:hparams}. The network is trained using the Adam optimizer with a learning rate of $2^{-8}$. The learning rate is decayed when the validation loss plateaus for 1,500 epochs. The default batch size is set to $2^{11}$ although many of the training data sizes are less than this value, so the batch size is automatically reduced to the size of the training data set when appropriate. The networks are trained for 8,192 epochs unless otherwise specified. The network is initialized using the Xavier uniform initialization scheme~\cite{glorotUnderstandingDifficultyTraining2010}, and the network activation function is GELU~\cite{hendrycksGaussianErrorLinear2020}. The final layer weights are initialized to zero, which heuristically led to faster convergence and better performance.

PINN-GM-III preprocesses its training data differently than PINN-GM-II. Explicitly, the position and potential are normalized by the characteristic length $x^{\star}$ and maximum potential $U^{\star}$ respectively. Using these characteristic scalars, a time constant can be computed and used in conjunction with $x^{\star}$ to non-dimensionalize the accelerations. This manifests through:
\begin{align}
    x = \frac{\bar{x}}{x^{\star}}, \quad U = \frac{\bar{U}}{U^{\star}}, \quad a = \frac{\bar{a}}{a^{\star}}
    \label{eq:non_dim}
\end{align}
where $x^{\star}$, $U^{\star}$, and $a^{\star}$ are the non-dimensionalization constants defined as:
\begin{align}
    x^{\star} = R, \quad U^{\star} = \max_i(\bar{U}_i - \bar{U}_{\text{LF}, i}) \label{eq:non_dim_U},  \quad t^{\star} = \sqrt{\frac{{x^{\star}}^2}{U^{\star}}},  \quad a^{\star} = \frac{x^{\star}}{{t^{\star}}^2}
\end{align}
where $R$ is the maximum radius of the celestial body, $\bar{U}_i$ is the true gravitational potential at the training datum at $\bm{x}_i$, and $U_{\text{LF}}$ is any low-fidelity potential contributions already accounted for within the PINN-GM (Section~\ref{sec:analytic_model}).

\begin{table}[htbp]
    \centering
    \begin{tabular}{lr} 
      \toprule
      \textbf{Hyperparameter} & \textbf{Value} \\
      \midrule
      learning\_rate          & $2^{-8}$          \\
      batch\_size             & $2^{11}$            \\
      num\_epochs             & $2^{13}$           \\
      optimizer               & Adam          \\
      activation\_function    & GELU          \\
      \bottomrule
    \end{tabular}
    \hspace{1cm}
    \begin{tabular}{lr} 
      \toprule
      \textbf{Hyperparameter} & \textbf{Value} \\
      \midrule
      lr\_scheduler               & plateau \\
      lr\_patience          & 1,500          \\
      decay\_rate             & 0.5            \\
      min\_delta             & 0.001           \\
      min\_lr               & 1e-6          \\
      \bottomrule
    \end{tabular}
    \caption{Default Set of Hyperparameters for Neural Network Training}
    \label{tab:hparams}
  \end{table}

\section{Hyperparameter Optimization}
\label{app:ablation}

Neural networks can be particularly sensitive to the correct choice of hyperparameters. This section seeks to characterize the sensitivity of the PINN-GM to these core hyperparameters, as well as determine the effect of different network sizes and quantity / quality of training data. Explicitly, network depth, width, batch size, learning rate, epochs, and loss function are varied, as well as the total amount data and its quality. As before, the heterogeneous density asteroid Eros detailed in Section~\ref{sec:case_study} is used to provide training data. 32,768 training data are sampled uniformly between 0-3R, and the network performance is evaluated using a mean percent error averaged across 4,096 separate validation samples for each of the proposed tests. 

\subsection{Network Size}
\begin{figure}[!htbp]
     \centering
     \begin{subfigure}[b]{0.49\textwidth}
          \centering
          \includegraphics[width=\textwidth]{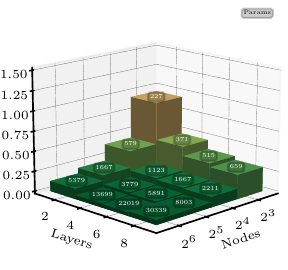} 
          \caption{Depth vs width}
          \label{fig:depth_width}
     \end{subfigure}
     \hfill
     \begin{subfigure}[b]{0.49\textwidth}
          \centering
          \includegraphics[width=\textwidth]{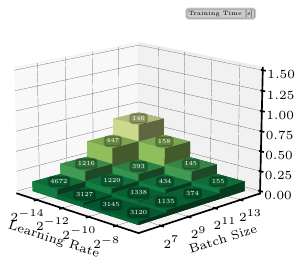} 
          \caption{Batch vs Learning Rate}
          \label{fig:batch_lr}
     \end{subfigure}
     \caption{Core Hyperparameters}
     \label{fig:core_hyperparameter_hists}
\end{figure}

The first test investigates the PINN-GM's sensitivity to network size, varying both network depth and width. The network depth is varied between ${2,4,6,8}$ hidden layers, and the width is varied between ${8, 16, 32, 64}$ nodes per hidden layer. The remaining hyperparameters are kept fixed at the defaults provided in Table~\ref{tab:hparams}. The mean percent error of the models are reported in Figure~\ref{fig:depth_width} with the corresponding model size, or total trainable parameters, overlaid on the bars.

Figure~\ref{fig:depth_width} demonstrates that PINN-GM-III performs well across a variety of different network sizes. The smallest networks will perform worse than the largest networks given their limited modeling capacity, yet despite this, all models remain below 5\% error. The smallest model of 227 trainable parameters averages at 0.9\% error, whereas the largest model with 30,339 parameters achieves 0.03\% error. This figure illustrates that optimal models require a minimum of 16 nodes with four hidden layers. The performance continues to improve with larger models, but only marginally. Therefore, a model of six hidden layers and 16 nodes per layer (1667 parameters) or six hidden layers and 32 nodes (5891 parameters) are recommended. 

\subsection{Batch Size and Learning Rate}

The next experiment studies the effect of batch size and learning rate on the PINN-GM. The network size is kept fixed at six hidden layers and 32 nodes per layer, but the batch size is varied between $\{2^7, 2^9, 2^{11}, 2^{13}\}$ and the learning rate is varied between $\{2^{-14},2^{-12}, 2^{-10},2^{-8}\}$. Again, the mean percent error of 4,096 samples are evaluated and shown in Figure~\ref{fig:batch_lr} with the total training time overlaid on the bars rather than parameter count. 

Figure~\ref{fig:batch_lr} demonstrates that there are a variety of learning rates and batch sizes that are acceptable for the PINN-GM. In general, smaller batch sizes produce more accurate models, albeit this comes with much longer training times. Regarding learning rate, higher is better. This is most likely coupled with the chosen learning rate scheduler, which decreases the learning rate by a factor of 0.5 every 1,500 epochs that the validation loss does not improve. Given this learning rate scheduler configuration, a learning rate of at least $2^{-8}$ is recommended for most applications. 

\subsection{Data Quantity and Epochs}
\begin{figure}[!htbp]
     \centering
     \begin{subfigure}[b]{0.49\textwidth}
          \centering
          \includegraphics[width=\textwidth]{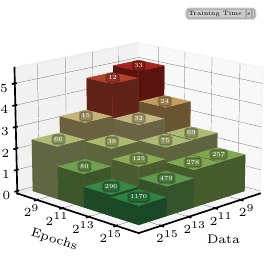} 
          \caption{Small Network (S)}
          \label{fig:small_data_epochs}
     \end{subfigure}
     \hfill
     \begin{subfigure}[b]{0.49\textwidth}
          \centering
          \includegraphics[width=\textwidth]{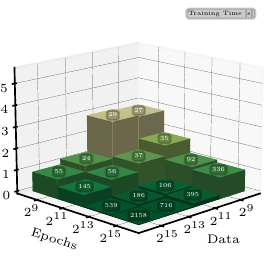} 
          \caption{Large Network (L)}
          \label{fig:large_data_epochs}
     \end{subfigure}
     \caption{Data vs Epochs}
     \label{fig:data_epochs}
\end{figure}

Fixing the learning rate to $2^{-8}$ and the batch size to $2^{11}$, a third experiment investigates the sensitivity of the PINN-GM to quantity of training data and length of training time. A PINN-III S and PINN-III L are prepared for this test. The PINN-III S uses the two layer, eight nodes per layer network configuration (227 parameters), whereas PINN-III L uses the eight layer, 64 node configuration (30,339 parameters). Both the PINN-III S and L are trained with increasing amounts of training data ranging from $\{2^9, 2^{11}, 2^{13}, 2^{15}\}$ samples and increasing the total number of training epochs from $\{2^9, 2^{11}, 2^{13}, 2^{15}\}$. As before, the mean percent error is evaluated and presented with the length of training time in Figures~\ref{fig:small_data_epochs} and \ref{fig:large_data_epochs}.

Figure~\ref{fig:small_data_epochs} shows that the PINN-III small is capable of achieving errors as low as 1\% given sufficient quantities of training data and training time. Performance does decrease as the number of samples and epochs decrease; however, all models converge to solutions with less than 5\% error. Similarly, Figure~\ref{fig:large_data_epochs} demonstrates that PINN-III L consistently benefits from additional training data and epochs, with the most accurate models reaching 0.04\% error. Unlike the PINN-III S, the L models always remained below 3\% error even in the low data and epoch regimes. Taken together, the results suggest that the PINN-GM-III should be trained for at minimum 8,192 epochs but can benefit from longer training if resources allow. 

\subsection{Data Quality and Physics Constraints}

The final experiment investigates the effect of additional physics constraints on model performance. Explicitly, an added Laplacian constraint in the network loss function improved model performance for the PINN-GM-II when trained on noisy data~\cite{martinSmallBodies2022}. This added constraint, however, also added considerable computational overhead to compute the second order derivative of the potential via automatic differentiation. This experiment seeks to determine if this term remains necessary given the new design modifications of the PINN-GM-III. 

\begin{figure}
     \centering
     \includegraphics[width=\textwidth]{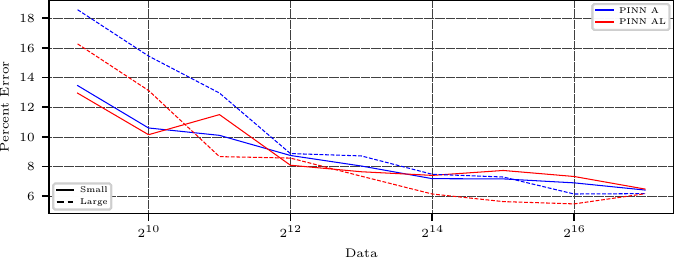}
     \caption{Error as a function of training data and loss function}
     \label{fig:noise_loss}
\end{figure}

This experiment begins by corrupting every acceleration vector by adding 10\% of their magnitude in a random direction to the truth vector as discussed in Section~\ref{sec:experiments}. The PINN-GM-III S and L are then trained on increasing amounts of this data ranging from $N=\{2^9, 2^{11}, 2^{13}, 2^{15}\}$. These models are trained once with the proposed loss function of Section~\ref{sec:loss_function} which only penalizes errors in the acceleration vector (PINN A) 
\begin{equation}
     \mathcal{L}_{\text{PINN A}} = \frac{1}{N}\sum^{N}_{i=0}\left(\left\lVert \minus \nabla \hat{U}(\bm{x}_i\vert \mathbf{\theta}) - \bm{a}_i\right\rVert + \frac{\lVert \minus \nabla \hat{U}(\bm{x}_i\vert \mathbf{\theta}) - \bm{a}_i\rVert }{\lVert \bm{a}_i\rVert }\right)
\end{equation}
and again with the Laplacian term added (PINN AL) or 
\begin{equation}
     \mathcal{L}_{\text{PINN AL}} = \frac{1}{N}\sum^{N}_{i=0}\left(\left\lVert \minus \nabla \hat{U}(\bm{x}_i\vert \mathbf{\theta}) - \bm{a}_i\right\rVert + \frac{\lVert \minus \nabla \hat{U}(\bm{x}_i\vert \mathbf{\theta}) - \bm{a}_i\rVert }{\lVert \bm{a}_i\rVert } + \left\lVert\nabla^2U(\bm{x}_i)\right\rVert\right)
\end{equation}
The batch size for this experiment is increased to $2^{15}$ to complete the experiment in a reasonable amount of training time, and the performance of the PINN-III L and S models as a function of training data are provided in Figure~\ref{fig:noise_loss}. 

Figure~\ref{fig:noise_loss} shows that the noise in the training data does deteriorate model performance; however even with low quantities of training data, the PINN-GM S is able to achieve errors as low as 13\%, just slightly above the noise floor. As the amount of training data increases, the S model reduces to approximately 7\% error. The fact that these models are capable of regressing solutions beneath the noise floor is a testament to the physics-informed nature of these models. By leveraging the known dynamics of the system, the models can better ignore parts of the training data that are fundamentally inconsistent with the physics. That said, the additional Laplacian component added to the loss function (PINN AL) does not yield a compelling advantage to warrant its added training time. Therefore, it is advised that the PINN-GM does not incorporate the additional Laplacian constraint in its loss function.

\section{Comments on Past Machine Learning Performance}
\label{app:ml_comparision}
Table \ref{tab:ML_Gravity} highlights the general performance of past and present machine learning gravity models. All values are taken from their corresponding reference, but further context is warranted as each model assessed accuracy in different ways and on different asteroids. This section aims to provide relevant details regarding these metrics for completeness. 

For the Gaussian process gravity model reported in Reference~\citenum{gaoEfficientGravityField2019}, the number of model parameters are not explicitly reported but can be deduced. Gaussian processes are defined by their covariance matrix and kernel function. The covariance matrix scales as $\mathcal{O}(N^2)$ and the maximum number of data points used were $N = 3,600$. This suggests that the minimum number of parameters used in the model is 12,960,000. The accuracy for these models are also reported at fixed radii from the center of mass for each asteroid rather than across the full domain, so the values can be considered upper-bounds. Moreover, the model is shown to diverge at high altitudes and therefore is not valid globally.

For the extreme learning machine gravity model reported in Reference~\citenum{furfaroModelingIrregularSmall2020}, the model size is determined by the fact that there are 50,000 hidden nodes in the ELM. The random weights connecting the three inputs to the 50,000 hidden nodes constitutes the first 150,000 parameters, and the weights connecting the hidden layer to the acceleration output correspond to the next 150,000 parameters. Each node in the hidden layer can also have a bias, adding another 50,000 points summing to total of 350,000 total model parameters. The asteroid modeled is 25143 Itokawa, and the error in Reference~\citenum{furfaroModelingIrregularSmall2020} is reported in terms of absolute terms rather than relative terms. Using Figure 9, it can be approximated that the relative error varies between 1\% and 10\%. 

In Reference~\citenum{chengRealtimeOptimalControl2020}, the neural network gravity model is reported to use 512 nodes per hidden layer for 6 hidden layers. The approximate number of weights and biases can be approximated by squaring the number of nodes per hidden layer and multiplying by the total number of layers minus one, and adding the biases for each node ($\approx512^2*(6-1) + 512*6 = 1,313,792$ parameters). The paper reports 1,000,000 training data were generated and divided into an 8:2 ratio between training and testing data, yielding 800,000 training data. The asteroid investigated is also 433-Eros and the average relative error of the test set is reported as 0.35\% in their Table 3. 

For GeodesyNets~\cite{izzoGeodesyIrregularSmall2022}, SIRENs of 9 hidden layers with 100 nodes each are used ($\approx 100^2 * (9-1) + 8*100= 80,800$). Four asteroids are studied: Bennu, Churyumov-Gerasimenko, Eros, and Itokawa. In their supplementary materials (Table S4), the relative error about Eros is reported at three characteristic altitudes. At their lowest altitude, the average error is 0.571\% and their highest altitude is 0.146\%. In attempts to quantify error across the entire high and low altitude regime, these values are averaged for the reported value of 0.359\%. The number of training data referenced is a result of the original paper sampling 1,000 data points every 10 iterations where the models were trained for 10,000 iterations, thereby equating to 1,000,000 training data. 

Finally for the PINN-GM-III, the average error reported for a network trained with 8 hidden layers with 16 nodes per layer. The model was trained on 4,096 data points distributed between 0-10R using the same hyperparameters specified in Appendix~\ref{app:ablation}, and achieved an average error on the validation set of 0.3\%. 

\section{Regression Details}
\label{app:regression_details}

In Section~\ref{sec:experiments}, multiple gravity models are regressed and tested on a variety of data conditions. The regression procedure for the neural networks is consistent with prior explanations; however, the remaining models require further context. This section aims to provide that context, explaining how each of the remaining gravity models are regressed on the data discussed in Section~\ref{sec:experiments}. 

\subsection{Spherical Harmonics}
Spherical harmonic gravity models are regressed by solving a linear system for their $p=N(N+1)$ total Stokes coefficients, where $N$ is the degree of the model. This system takes the following form:
\begin{align}
    \bm{a} &= H \bm{c} \\
    H^T \bm{a} &= H^T H \bm{c} \\
    (H^T H)^{-1} H^T \bm{a} &= \bm{c} 
\end{align}
where $\bm{c}$ are the vector of spherical harmonic coefficients $C_{lm}$ and $S_{lm}$, $\bm{a}$ are the vector of accelerations, and $H$ is the jacobian $\frac{\partial \bm{a}}{\partial \bm{c}}\vert_{\bm{r}}$, and $\bm{r}$ is the position vector for each test point. Direct least square solutions may be appropriate for small systems, but when many harmonics need to be regressed and/or the amount of data used to regress them increases, least squares becomes computationally infeasible. Moreover, high-degree spherical harmonic models pose numerical instabilities as the observability of the harmonics decays rapidly as $(R/r)^l$, often making $H^T H$ ill-conditioned. 

Kaula's rule --- a form of ridge regression that is used to regularize the spherical harmonic coefficients --- can help eliminate the ill-conditioned nature of the high-degree regression. Rather than seeking to minimize the mean squared error solution 
\begin{equation}
    L(\bm{c}) = \lVert \bm{a} - H\bm{c} \rVert^2
\end{equation}
ridge regression adds a penalty term to ensure the coefficients decay in magnitude for higher degree harmonics through
\begin{equation}
    L(\bm{c}) = \lVert \bm{a} - H\bm{c} \rVert^2 + \lVert \Gamma \bm{c} \rVert^2
\end{equation}
where $\Gamma$ are the regularization matrix defined through Kaula's rule 
\begin{equation}
    \Gamma_{ll} = \begin{cases}
        \frac{\alpha}{l^2} & l > 0 \\
        1 & l = 0
    \end{cases}
\end{equation}
where $\alpha$ is a user specified constant typically chosen through cross validation. The corresponding solution to the ridge regression then becomes 
\begin{equation}
    (H^T H + \Gamma)^{-1} H^T \bm{a} = \bm{c} \\
\end{equation}
which yields a solution with increasingly small spherical harmonic coefficients at high-degree. This simultaneously removes the ill-posed nature of the original regression and also provides a framework for supporting low-data regression. 

While using Kaula's rule mitigates the ill-posedness of the original regression, it remains computationally expensive to invert $(H^T H + \Gamma)$ for large datasets. This is solved with two strategies: 1) recursive least squares and 2) iterative coefficient regression. Recursive least square sequentially feeds in small batches of data to maintain computational tractability, while iterative coefficient regression regresses low-degree harmonics before the high-degree harmonics.

For the regression used in Section~\ref{sec:experiments}, recursive least squares is performed in batches of 100 position / acceleration pairs using the following recursion relationships
\begin{align}
    K_{i+1}^{-1} &= K_{i}^{-1} - K_{i}^{-1}  H_k^T  (I + H_k  K_{i}^{-1}  H_k^T)^{-1}  H_k  K_{i}^{-1}\\
    \bm{c}_{k+1} &= \bm{c}_k + K_{i+1}^{-1}  H_k^T  (\bm{a}_k - H_k \bm{c}_k)
    \label{eq:RLLS_K}
\end{align}
where $K_0^{-1} = (H_0^T H_0 + \Gamma)^{-1}$, and the iterative coefficient regression is performed by only regressing 5,000 coefficients at a time. In addition, samples beneath the Brillouin surface $r < R$ are purposefully omitted from the regression due to the $(R/r)^l$ scaling in the harmonic model. For very high degree models, this term diverges when evaluated on sub-Brillouin samples, which breaks the regression. 

\subsection{Mascons}

Like spherical harmonics, mascons regression is also a form of least squares regression. Specifically, a set of $N$ mascons are uniformly distributed through the volume of the shape model, and their corresponding masses are regressed. Ideally, the mascon regression should also include a non-linear constraint applied to all mass elements that ensures they always have a value greater than zero to ensure compliance with physics --- i.e. there are no such things as negative masses. This physical compliance, however, often comes at a major loss of modeling accuracy. Therefore, Section~\ref{sec:experiments} allows for negative masses to be regressed to give the model the strongest cases against the PINN-GM. 

In addition to allowing negative masses, the mascon regression also requires iterative fitting when the model sizes are large and there exists much data. To do this, the mascons are distributed and fit in batches of 500 using all available data. After each batch is fit, their contribution to the acceleration vectors are removed, the next batch is randomly distributed throughout the volume, and the model fits to the acceleration residuals. This process is repeated until the total mass elements are reached. 

The total number of parameters for a mascon model correspond with four parameters per mascon, corresponding to the three-component position of the mass, $\bm{r}_k$, and it's associated gravitational parameter $\mu_k$. Therefore the total parameter count for the mascon model is $p = 4N$.

\subsection{Extreme Learning Machines}

Extreme learning machines regression also closely resembles that of the spherical harmonic regression in that it also uses ridge regression and recursive least squares optimization. Explicitly, the ELM models are regressed by applying a fixed ridge regression matrix $\Gamma_{\text{ELM}} = \alpha \mathbb{I}$ and breaking the dataset into batches to be applied recursively using Eqs.~\ref{eq:RLLS_K}. Notably, the ELM also relies on a random non-linear projection into a higher-dimensional space before the linear regression as described in detail in \cite{furfaroModelingIrregularSmall2020}. For the comparison study in this paper, all input and output data are preprocessed using a min-max transformation to $[0,1]$ and applying a sigmoid activation function at the hidden nodes. The total parameter count for these models can be computed by summing the weights from the three inputs to the hidden layer, the biases of the hidden layer, and the weights to the three outputs, totaling to: $3N + N + 3N = 7N$. 

\subsection{Polyhedral Models}
The regression of polyhedral models follows an entirely unrelated process to the other gravity models, requiring image data to construct the shape model of the body using stereophotoclinometry. Given that these shape models cannot be computed directly from position and acceleration data, Section~\ref{sec:experiments} instead only includes the performance of polyhedral models that are of similar parameter counts to the other models and assumes these models are perfectly regressed. The model sizes of the polyhedral models are computed by summing the positions of each vertex in the model, and the indices (stored as long integers) which identify the vertices that comprise each face. Therefore the total model size can be computed via $p=3V + \frac{3F}{2}$ where $F$ corresponds to the number of facets and $V$ corresponds to the number of vertices. The $3F/2$ captures the fact that long integers are half the size of 32-bit floats.  

\subsection{Neural Network Models}
The regression of the neural network models is detailed at the beginning of Section~\ref{sec:pinn_gm}. The parameter count for these models corresponds with the weights connecting the $l - 1$ hidden layers, the weights connecting the inputs to the first hidden layer, the weights connecting the last hidden layer to the outputs, and the biases for all nodes. Therefore, the total parameters can be estimated with $3N + N^2(l-1) + N + lN$ respectively where $N$ is the width of the network and $l$ is the number of hidden layers. Note that the traditional neural networks will have three outputs, whereas PINNs only have one.

\subsection{GeodesyNet}
The regression of the GeodesyNets follows the direct training method outlined in Ref.~\cite{izzoGeodesyIrregularSmall2022}. The direct training method is chosen over the differential training method because the former does not assume knowledge of the asteroid shape --- therefore matching the data conditions for the other machine learning gravity models. 

For their regression, the GeodesyNets use 300,000 quadrature points to numerically evaluate the acceleration at a field point. In the original paper, the models are trained for 10,000 iterations, where 1,000 new training data are randomly generated in the unit volume --- i.e. near the body --- once every 10 iterations. In this manuscript, the data is generated a priori --- either 500 or 50,000 points spanning 0-10R --- and 1,000 data points are randomly sampled from this generated data once every 10 iterations to mimic the original training configuration. The architecture of the GeodesyNet matched that of the other machine learning models. The small GeodesyNet had four hidden layers and eight nodes per hidden layer, and the large GeodesyNet had eight hidden layers with 64 nodes per hidden layer. All training data were non-dimensionalized to ensure that all of the asteroid density remains within the unit cube.

The performance of the GeodesyNets in the comparative study is believed to be the result of the small dataset size and its distribution across the training volume. As discussed, the original GeodesyNet training paper achieves high accuracy by training on 1,000,000 data that exist within the unit cube or sphere; however, the data used in the comparative study experiments span a spherical volume out to 10R. This yields a volume that is approximately 1,000 times larger (volume scales cubically) with considerably less data occupying that volume. As such, the GeodesyNets do not have enough data to regress an accurate density field across the entire unit volume.

\subsection{Supplemental Comparison Study Results}
\label{app:comparison_study}

The following tables report the exact metric values used for the comparison study, as well as the associated rank across all models. 

\newpage 

\clearpage
\subsubsection{Eros Heterogeneous Density}

\begin{table}[h!]
\centering
\scalebox{0.6}{
\begin{tabular}{rlllllllll}
\toprule
{} &                                      N &            \makecell{Error \\ (\%)} &                                Score &               \makecell{Planes \\ (\%)} &             \makecell{Interior \\ (\%)} &             \makecell{Exterior \\ (\%)} &              \makecell{Extrap. \\ (\%)} &               \makecell{Surface \\ (\%)} &                \makecell{Traj. \\ (km)} \\
Model               &                                        &                                     &                                      &                                         &                                         &                                         &                                         &                                          &                                         \\
\midrule
MASCONS L           &  \cellcolor[rgb]{1.00,1.00,1.00} 50000 &   \cellcolor[rgb]{1.00,1.00,1.00} 0 &   \cellcolor[rgb]{0.37,0.31,0.64} 14 &  \cellcolor[rgb]{0.37,0.31,0.64} 1.2E-2 &    \cellcolor[rgb]{0.31,0.38,0.67} 0.13 &  \cellcolor[rgb]{0.34,0.34,0.65} 5.4E-4 &  \cellcolor[rgb]{0.34,0.34,0.65} 3.4E-4 &      \cellcolor[rgb]{0.28,0.42,0.69} 8.3 &  \cellcolor[rgb]{0.35,0.34,0.65} 1.7E-3 \\
\textbf{PINN III S} &  \cellcolor[rgb]{1.00,1.00,1.00} 50000 &   \cellcolor[rgb]{1.00,1.00,1.00} 0 &   \cellcolor[rgb]{0.35,0.34,0.65} 29 &    \cellcolor[rgb]{0.29,0.41,0.69} 0.17 &     \cellcolor[rgb]{0.25,0.46,0.71} 1.2 &  \cellcolor[rgb]{0.29,0.41,0.69} 2.8E-2 &   \cellcolor[rgb]{0.31,0.39,0.67} 0.065 &       \cellcolor[rgb]{0.24,0.58,0.72} 11 &  \cellcolor[rgb]{0.27,0.44,0.70} 4.2E-2 \\
\textbf{PINN III L} &  \cellcolor[rgb]{1.00,1.00,1.00} 50000 &   \cellcolor[rgb]{1.00,1.00,1.00} 0 &   \cellcolor[rgb]{0.32,0.37,0.66} 34 &  \cellcolor[rgb]{0.34,0.34,0.65} 1.3E-2 &   \cellcolor[rgb]{0.34,0.34,0.65} 0.099 &  \cellcolor[rgb]{0.31,0.38,0.67} 1.7E-3 &     \cellcolor[rgb]{0.89,0.96,0.60} 1.3 &      \cellcolor[rgb]{0.34,0.34,0.65} 3.2 &  \cellcolor[rgb]{0.32,0.37,0.66} 2.7E-3 \\
\bottomrule
\end{tabular}

    }
    \caption{Metric values for each gravity model trained on heterogeneous Eros gravity field data.}
\end{table}

\begin{table}[h!]
\centering
\scalebox{0.6}{
    \begin{tabular}{rlllllllll}
\toprule
{} &                                      N &            \makecell{Error \\ (\%)} &          \makecell{Score \\ (Rank)} &          \makecell{Planes \\ (Rank)} &        \makecell{Interior \\ (Rank)} &        \makecell{Exterior \\ (Rank)} &         \makecell{Extrap. \\ (Rank)} &         \makecell{Surface \\ (Rank)} &           \makecell{Traj. \\ (Rank)} \\
Model               &                                        &                                     &                                     &                                      &                                      &                                      &                                      &                                      &                                      \\
\midrule
MASCONS L           &  \cellcolor[rgb]{1.00,1.00,1.00} 50000 &   \cellcolor[rgb]{1.00,1.00,1.00} 0 &   \cellcolor[rgb]{0.37,0.31,0.64} 1 &  \cellcolor[rgb]{0.37,0.31,0.64} 1.0 &  \cellcolor[rgb]{0.31,0.38,0.67} 3.0 &  \cellcolor[rgb]{0.34,0.34,0.65} 2.0 &  \cellcolor[rgb]{0.34,0.34,0.65} 2.0 &  \cellcolor[rgb]{0.28,0.42,0.69} 4.0 &  \cellcolor[rgb]{0.35,0.34,0.65} 2.0 \\
\textbf{PINN III S} &  \cellcolor[rgb]{1.00,1.00,1.00} 50000 &   \cellcolor[rgb]{1.00,1.00,1.00} 0 &   \cellcolor[rgb]{0.35,0.34,0.65} 2 &  \cellcolor[rgb]{0.29,0.41,0.69} 4.0 &  \cellcolor[rgb]{0.25,0.46,0.71} 5.0 &  \cellcolor[rgb]{0.29,0.41,0.69} 4.0 &  \cellcolor[rgb]{0.31,0.39,0.67} 3.0 &  \cellcolor[rgb]{0.24,0.58,0.72} 8.0 &  \cellcolor[rgb]{0.27,0.44,0.70} 5.0 \\
\textbf{PINN III L} &  \cellcolor[rgb]{1.00,1.00,1.00} 50000 &   \cellcolor[rgb]{1.00,1.00,1.00} 0 &   \cellcolor[rgb]{0.32,0.37,0.66} 3 &  \cellcolor[rgb]{0.34,0.34,0.65} 2.0 &  \cellcolor[rgb]{0.34,0.34,0.65} 2.0 &  \cellcolor[rgb]{0.31,0.38,0.67} 3.0 &   \cellcolor[rgb]{0.89,0.96,0.60} 22 &  \cellcolor[rgb]{0.34,0.34,0.65} 2.0 &  \cellcolor[rgb]{0.32,0.37,0.66} 3.0 \\
\bottomrule
\end{tabular}

    }
    \caption{Rank values for each gravity model trained on heterogeneous Eros gravity field data.}
    \label{tab:eros_rank_comparison_study}
\end{table}

\clearpage
\subsubsection{Eros Homogeneous Density}

\begin{table}[h!]
\centering
\scalebox{0.6}{
    \begin{tabular}{rlllllllll}
\toprule
{} &                                      N &            \makecell{Error \\ (\%)} &                                Score &               \makecell{Planes \\ (\%)} &             \makecell{Interior \\ (\%)} &             \makecell{Exterior \\ (\%)} &              \makecell{Extrap. \\ (\%)} &               \makecell{Surface \\ (\%)} &                \makecell{Traj. \\ (km)} \\
Model               &                                        &                                     &                                      &                                         &                                         &                                         &                                         &                                          &                                         \\
\midrule
POLY L              &       \cellcolor[rgb]{0.59,0.59,0.59}  &    \cellcolor[rgb]{0.59,0.59,0.59}  &   \cellcolor[rgb]{0.37,0.31,0.64} 11 &  \cellcolor[rgb]{0.37,0.31,0.64} 2.7E-3 &  \cellcolor[rgb]{0.37,0.31,0.64} 1.2E-2 &  \cellcolor[rgb]{0.31,0.38,0.67} 8.9E-4 &  \cellcolor[rgb]{0.34,0.34,0.65} 3.1E-5 &    \cellcolor[rgb]{0.37,0.31,0.64} 0.090 &  \cellcolor[rgb]{0.32,0.37,0.66} 2.2E-3 \\
MASCONS L           &  \cellcolor[rgb]{1.00,1.00,1.00} 50000 &   \cellcolor[rgb]{1.00,1.00,1.00} 0 &   \cellcolor[rgb]{0.35,0.34,0.65} 20 &  \cellcolor[rgb]{0.34,0.34,0.65} 1.2E-2 &    \cellcolor[rgb]{0.31,0.38,0.67} 0.11 &  \cellcolor[rgb]{0.34,0.34,0.65} 5.2E-4 &  \cellcolor[rgb]{0.31,0.39,0.67} 3.4E-4 &      \cellcolor[rgb]{0.24,0.58,0.72} 8.0 &  \cellcolor[rgb]{0.35,0.34,0.65} 1.5E-3 \\
\textbf{PINN III L} &  \cellcolor[rgb]{1.00,1.00,1.00} 50000 &   \cellcolor[rgb]{1.00,1.00,1.00} 0 &   \cellcolor[rgb]{0.32,0.37,0.66} 37 &  \cellcolor[rgb]{0.31,0.38,0.67} 2.1E-2 &    \cellcolor[rgb]{0.28,0.42,0.69} 0.20 &  \cellcolor[rgb]{0.28,0.42,0.69} 3.5E-3 &    \cellcolor[rgb]{0.76,0.90,0.63} 0.52 &      \cellcolor[rgb]{0.31,0.38,0.67} 3.7 &  \cellcolor[rgb]{0.29,0.41,0.68} 3.2E-2 \\
\bottomrule
\end{tabular}

    }
    \caption{Metric values for each gravity model trained on homogeneous Eros gravity field data.}
\end{table}

\begin{table}[h!]
\centering
\scalebox{0.6}{
    \begin{tabular}{rlllllllll}
\toprule
{} &                                      N &            \makecell{Error \\ (\%)} &          \makecell{Score \\ (Rank)} &          \makecell{Planes \\ (Rank)} &        \makecell{Interior \\ (Rank)} &        \makecell{Exterior \\ (Rank)} &         \makecell{Extrap. \\ (Rank)} &         \makecell{Surface \\ (Rank)} &           \makecell{Traj. \\ (Rank)} \\
Model               &                                        &                                     &                                     &                                      &                                      &                                      &                                      &                                      &                                      \\
\midrule
POLY L              &       \cellcolor[rgb]{0.59,0.59,0.59}  &    \cellcolor[rgb]{0.59,0.59,0.59}  &   \cellcolor[rgb]{0.37,0.31,0.64} 1 &  \cellcolor[rgb]{0.37,0.31,0.64} 1.0 &  \cellcolor[rgb]{0.37,0.31,0.64} 1.0 &  \cellcolor[rgb]{0.31,0.38,0.67} 3.0 &  \cellcolor[rgb]{0.34,0.34,0.65} 2.0 &  \cellcolor[rgb]{0.37,0.31,0.64} 1.0 &  \cellcolor[rgb]{0.32,0.37,0.66} 3.0 \\
MASCONS L           &  \cellcolor[rgb]{1.00,1.00,1.00} 50000 &   \cellcolor[rgb]{1.00,1.00,1.00} 0 &   \cellcolor[rgb]{0.35,0.34,0.65} 2 &  \cellcolor[rgb]{0.34,0.34,0.65} 2.0 &  \cellcolor[rgb]{0.31,0.38,0.67} 3.0 &  \cellcolor[rgb]{0.34,0.34,0.65} 2.0 &  \cellcolor[rgb]{0.31,0.39,0.67} 3.0 &  \cellcolor[rgb]{0.24,0.58,0.72} 8.0 &  \cellcolor[rgb]{0.35,0.34,0.65} 2.0 \\
\textbf{PINN III L} &  \cellcolor[rgb]{1.00,1.00,1.00} 50000 &   \cellcolor[rgb]{1.00,1.00,1.00} 0 &   \cellcolor[rgb]{0.32,0.37,0.66} 3 &  \cellcolor[rgb]{0.31,0.38,0.67} 3.0 &  \cellcolor[rgb]{0.28,0.42,0.69} 4.0 &  \cellcolor[rgb]{0.28,0.42,0.69} 4.0 &   \cellcolor[rgb]{0.76,0.90,0.63} 19 &  \cellcolor[rgb]{0.31,0.38,0.67} 3.0 &  \cellcolor[rgb]{0.29,0.41,0.68} 4.0 \\
\bottomrule
\end{tabular}

    }
    \caption{Rank values for each gravity model trained on homogeneous Eros gravity field data.}
\end{table}

\clearpage
\subsubsection{Bennu Heterogeneous Density}

\begin{table}[h!]
\centering
\scalebox{0.6}{
    \begin{tabular}{rlllllllll}
\toprule
{} &                                      N &            \makecell{Error \\ (\%)} &                                Score &               \makecell{Planes \\ (\%)} &             \makecell{Interior \\ (\%)} &             \makecell{Exterior \\ (\%)} &              \makecell{Extrap. \\ (\%)} &              \makecell{Surface \\ (\%)} &                 \makecell{Traj. \\ (km)} \\
Model               &                                        &                                     &                                      &                                         &                                         &                                         &                                         &                                         &                                          \\
\midrule
MASCONS L           &  \cellcolor[rgb]{1.00,1.00,1.00} 50000 &   \cellcolor[rgb]{1.00,1.00,1.00} 0 &   \cellcolor[rgb]{0.37,0.31,0.64} 13 &  \cellcolor[rgb]{0.37,0.31,0.64} 1.0E-2 &    \cellcolor[rgb]{0.37,0.31,0.64} 0.37 &  \cellcolor[rgb]{0.37,0.31,0.64} 1.4E-3 &  \cellcolor[rgb]{0.34,0.34,0.65} 7.1E-4 &     \cellcolor[rgb]{0.25,0.46,0.71} 3.3 &   \cellcolor[rgb]{0.32,0.37,0.66} 5.4E-5 \\
\textbf{PINN III L} &  \cellcolor[rgb]{1.00,1.00,1.00} 50000 &   \cellcolor[rgb]{1.00,1.00,1.00} 0 &   \cellcolor[rgb]{0.35,0.34,0.65} 26 &  \cellcolor[rgb]{0.34,0.34,0.65} 1.4E-2 &    \cellcolor[rgb]{0.31,0.38,0.67} 0.62 &  \cellcolor[rgb]{0.34,0.34,0.65} 1.6E-3 &    \cellcolor[rgb]{0.46,0.79,0.65} 0.31 &     \cellcolor[rgb]{0.28,0.42,0.69} 3.0 &   \cellcolor[rgb]{0.35,0.34,0.65} 4.4E-5 \\
\textbf{PINN III S} &  \cellcolor[rgb]{1.00,1.00,1.00} 50000 &   \cellcolor[rgb]{1.00,1.00,1.00} 0 &   \cellcolor[rgb]{0.32,0.37,0.66} 31 &    \cellcolor[rgb]{0.29,0.41,0.69} 0.11 &     \cellcolor[rgb]{0.25,0.46,0.71} 1.6 &  \cellcolor[rgb]{0.29,0.41,0.69} 3.1E-2 &  \cellcolor[rgb]{0.24,0.48,0.71} 3.2E-2 &     \cellcolor[rgb]{0.23,0.49,0.72} 3.3 &   \cellcolor[rgb]{0.22,0.50,0.73} 1.9E-4 \\
\bottomrule
\end{tabular}

    }
    \caption{Metric values for each gravity model trained on heterogeneous Bennu gravity field data.}
\end{table}

\begin{table}[h!]
\centering
\scalebox{0.6}{
    \begin{tabular}{rlllllllll}
\toprule
{} &                                      N &            \makecell{Error \\ (\%)} &          \makecell{Score \\ (Rank)} &          \makecell{Planes \\ (Rank)} &        \makecell{Interior \\ (Rank)} &        \makecell{Exterior \\ (Rank)} &         \makecell{Extrap. \\ (Rank)} &         \makecell{Surface \\ (Rank)} &           \makecell{Traj. \\ (Rank)} \\
Model               &                                        &                                     &                                     &                                      &                                      &                                      &                                      &                                      &                                      \\
\midrule
MASCONS L           &  \cellcolor[rgb]{1.00,1.00,1.00} 50000 &   \cellcolor[rgb]{1.00,1.00,1.00} 0 &   \cellcolor[rgb]{0.37,0.31,0.64} 1 &  \cellcolor[rgb]{0.37,0.31,0.64} 1.0 &  \cellcolor[rgb]{0.37,0.31,0.64} 1.0 &  \cellcolor[rgb]{0.37,0.31,0.64} 1.0 &  \cellcolor[rgb]{0.34,0.34,0.65} 2.0 &  \cellcolor[rgb]{0.25,0.46,0.71} 5.0 &  \cellcolor[rgb]{0.32,0.37,0.66} 3.0 \\
\textbf{PINN III L} &  \cellcolor[rgb]{1.00,1.00,1.00} 50000 &   \cellcolor[rgb]{1.00,1.00,1.00} 0 &   \cellcolor[rgb]{0.35,0.34,0.65} 2 &  \cellcolor[rgb]{0.34,0.34,0.65} 2.0 &  \cellcolor[rgb]{0.31,0.38,0.67} 3.0 &  \cellcolor[rgb]{0.34,0.34,0.65} 2.0 &   \cellcolor[rgb]{0.46,0.79,0.65} 13 &  \cellcolor[rgb]{0.28,0.42,0.69} 4.0 &  \cellcolor[rgb]{0.35,0.34,0.65} 2.0 \\
\textbf{PINN III S} &  \cellcolor[rgb]{1.00,1.00,1.00} 50000 &   \cellcolor[rgb]{1.00,1.00,1.00} 0 &   \cellcolor[rgb]{0.32,0.37,0.66} 3 &  \cellcolor[rgb]{0.29,0.41,0.69} 4.0 &  \cellcolor[rgb]{0.25,0.46,0.71} 5.0 &  \cellcolor[rgb]{0.29,0.41,0.69} 4.0 &  \cellcolor[rgb]{0.24,0.48,0.71} 5.0 &  \cellcolor[rgb]{0.23,0.49,0.72} 6.0 &  \cellcolor[rgb]{0.22,0.50,0.73} 7.0 \\
\bottomrule
\end{tabular}

    }
    \caption{Rank values for each gravity model trained on heterogeneous Bennu gravity field data.}
\end{table}

\clearpage
\subsubsection{Bennu Homogeneous Density}

\begin{table}[h!]
\centering
\scalebox{0.6}{
    \begin{tabular}{rlllllllll}
\toprule
{} &                                      N &            \makecell{Error \\ (\%)} &                                Score &               \makecell{Planes \\ (\%)} &             \makecell{Interior \\ (\%)} &             \makecell{Exterior \\ (\%)} &              \makecell{Extrap. \\ (\%)} &              \makecell{Surface \\ (\%)} &                 \makecell{Traj. \\ (km)} \\
Model               &                                        &                                     &                                      &                                         &                                         &                                         &                                         &                                         &                                          \\
\midrule
POLY L              &       \cellcolor[rgb]{0.59,0.59,0.59}  &    \cellcolor[rgb]{0.59,0.59,0.59}  &    \cellcolor[rgb]{0.37,0.31,0.64} 8 &  \cellcolor[rgb]{0.37,0.31,0.64} 2.9E-3 &  \cellcolor[rgb]{0.37,0.31,0.64} 3.6E-2 &  \cellcolor[rgb]{0.37,0.31,0.64} 1.0E-3 &  \cellcolor[rgb]{0.34,0.34,0.65} 6.9E-5 &    \cellcolor[rgb]{0.37,0.31,0.64} 0.21 &   \cellcolor[rgb]{0.35,0.34,0.65} 1.1E-5 \\
MASCONS L           &  \cellcolor[rgb]{1.00,1.00,1.00} 50000 &   \cellcolor[rgb]{1.00,1.00,1.00} 0 &   \cellcolor[rgb]{0.35,0.34,0.65} 19 &  \cellcolor[rgb]{0.34,0.34,0.65} 9.8E-3 &    \cellcolor[rgb]{0.34,0.34,0.65} 0.34 &  \cellcolor[rgb]{0.34,0.34,0.65} 1.5E-3 &  \cellcolor[rgb]{0.31,0.39,0.67} 7.1E-4 &     \cellcolor[rgb]{0.20,0.54,0.74} 3.2 &   \cellcolor[rgb]{0.32,0.37,0.66} 5.3E-5 \\
\textbf{PINN III S} &  \cellcolor[rgb]{1.00,1.00,1.00} 50000 &   \cellcolor[rgb]{1.00,1.00,1.00} 0 &   \cellcolor[rgb]{0.29,0.41,0.68} 35 &    \cellcolor[rgb]{0.26,0.45,0.70} 0.11 &     \cellcolor[rgb]{0.20,0.53,0.74} 1.4 &  \cellcolor[rgb]{0.25,0.46,0.71} 3.1E-2 &  \cellcolor[rgb]{0.21,0.52,0.73} 1.1E-2 &     \cellcolor[rgb]{0.25,0.46,0.71} 2.8 &   \cellcolor[rgb]{0.22,0.50,0.73} 3.7E-4 \\
\bottomrule
\end{tabular}

    }
    \caption{Metric values for each gravity model trained on homogeneous Bennu gravity field data.}
\end{table}

\begin{table}[b!]
\centering
\scalebox{0.6}{
    \begin{tabular}{rlllllllll}
\toprule
{} &                                      N &            \makecell{Error \\ (\%)} &          \makecell{Score \\ (Rank)} &          \makecell{Planes \\ (Rank)} &        \makecell{Interior \\ (Rank)} &        \makecell{Exterior \\ (Rank)} &         \makecell{Extrap. \\ (Rank)} &         \makecell{Surface \\ (Rank)} &           \makecell{Traj. \\ (Rank)} \\
Model               &                                        &                                     &                                     &                                      &                                      &                                      &                                      &                                      &                                      \\
\midrule
POLY L              &       \cellcolor[rgb]{0.59,0.59,0.59}  &    \cellcolor[rgb]{0.59,0.59,0.59}  &   \cellcolor[rgb]{0.37,0.31,0.64} 1 &  \cellcolor[rgb]{0.37,0.31,0.64} 1.0 &  \cellcolor[rgb]{0.37,0.31,0.64} 1.0 &  \cellcolor[rgb]{0.37,0.31,0.64} 1.0 &  \cellcolor[rgb]{0.34,0.34,0.65} 2.0 &  \cellcolor[rgb]{0.37,0.31,0.64} 1.0 &  \cellcolor[rgb]{0.35,0.34,0.65} 2.0 \\
MASCONS L           &  \cellcolor[rgb]{1.00,1.00,1.00} 50000 &   \cellcolor[rgb]{1.00,1.00,1.00} 0 &   \cellcolor[rgb]{0.35,0.34,0.65} 2 &  \cellcolor[rgb]{0.34,0.34,0.65} 2.0 &  \cellcolor[rgb]{0.34,0.34,0.65} 2.0 &  \cellcolor[rgb]{0.34,0.34,0.65} 2.0 &  \cellcolor[rgb]{0.31,0.39,0.67} 3.0 &  \cellcolor[rgb]{0.20,0.54,0.74} 7.0 &  \cellcolor[rgb]{0.32,0.37,0.66} 3.0 \\
\textbf{PINN III S} &  \cellcolor[rgb]{1.00,1.00,1.00} 50000 &   \cellcolor[rgb]{1.00,1.00,1.00} 0 &   \cellcolor[rgb]{0.29,0.41,0.68} 4 &  \cellcolor[rgb]{0.26,0.45,0.70} 5.0 &  \cellcolor[rgb]{0.20,0.53,0.74} 7.0 &  \cellcolor[rgb]{0.25,0.46,0.71} 5.0 &  \cellcolor[rgb]{0.21,0.52,0.73} 6.0 &  \cellcolor[rgb]{0.25,0.46,0.71} 5.0 &  \cellcolor[rgb]{0.22,0.50,0.73} 7.0 \\
\bottomrule
\end{tabular}

    }
    \caption{Rank values for each gravity model trained on homogeneous Bennu gravity field data.}
\end{table}

\end{document}